\documentclass[sigconf]{acmart}

\AtBeginDocument{%
  \providecommand\BibTeX{{%
    \normalfont B\kern-0.5em{\scshape i\kern-0.25em b}\kern-0.8em\TeX}}}

\copyrightyear{2021} 
\acmYear{2021} 
\setcopyright{acmcopyright}\acmConference[MM '21]{Proceedings of the 29th ACM International Conference on Multimedia}{October 20--24, 2021}{Virtual Event, China}
\acmBooktitle{Proceedings of the 29th ACM International Conference on Multimedia (MM '21), October 20--24, 2021, Virtual Event, China}
\acmPrice{15.00}
\acmDOI{10.1145/3474085.3475372}
\acmISBN{978-1-4503-8651-7/21/10}

\usepackage{graphicx}
\usepackage{amsmath}
\usepackage{subfigure}
\usepackage{multirow}
\usepackage{color}
\usepackage{balance}

\acmSubmissionID{1218}
\begin{document}
\fancyhead{}
\title{ASFD: Automatic and Scalable Face Detector}


\author{Jian Li}
\authornote{Both authors contributed equally to this research.}
\email{swordli@tencent.com}
\affiliation{%
  \institution{Tencent Youtu Lab}
  \city{Shanghai}
  \country{China}
}

\author{Bin Zhang}
\authornotemark[1]
\email{z-bingo@seu.edu.cn}
\affiliation{%
  \institution{Southeast University}
  \city{Nanjing}
  \country{China}
}

\author{Yabiao Wang}
\email{caseywang@tencent.com}
\affiliation{%
  \institution{Tencent Youtu Lab}
  \city{Shanghai}
  \country{China}
}

\author{Ying Tai}
\email{yingtai@tencent.com}
\affiliation{%
  \institution{Tencent Youtu Lab}
  \city{Shanghai}
  \country{China}
}

\author{Zhenyu Zhang}
\email{joeyzyzhang@tencent.com}
\affiliation{%
  \institution{Tencent Youtu Lab}
  \city{Shanghai}
  \country{China}
}

\author{Chengjie Wang}
\email{jasoncjwang@tencent.com}
\affiliation{%
  \institution{Tencent Youtu Lab}
  \city{Shanghai}
  \country{China}
}

\author{Jilin Li}
\email{jerolinli@tencent.com}
\affiliation{%
  \institution{Tencent Youtu Lab}
  \city{Shanghai}
  \country{China}
}

\author{Xiaoming Huang}
\email{skyhuang@tencent.com}
\affiliation{%
  \institution{Tencent Youtu Lab}
  \city{Shanghai}
  \country{China}
}

\author{Yili Xia}
\authornote{Correspondence author.}
\email{yili_xia@seu.edu.cn}
\affiliation{%
  \institution{Southeast University}
  \city{Nanjing}
  \country{China}
}

%\author{Ying Tai}
%\affiliation{%
%  \institution{Youtu Lab, Tencent}
%  \state{Shanghai}
%  \country{China}
% }
%\email{yingtai@tencent.com}

%\author{Chengjie Wang}
%\affiliation{%
%  \institution{Youtu Lab, Tencent}
%  \state{Shanghai}
%  \country{China}
% }
%\email{jasoncjwang@tencent.com}

%\author{Jilin Li}
%\affiliation{%
%  \institution{Youtu Lab, Tencent}
%  \state{Shanghai}
%  \country{China}
% }
%\email{ jerolinli@tencent.com}

%\author{Xiaoming Huang}
%\affiliation{%
%  \institution{Youtu Lab, Tencent}
%  \state{Shanghai}
%  \country{China}
% }
%\email{skyhuang@tencent.com}

%\author{Yili Xia}
%\affiliation{%
%  \institution{Southeast University}
%  \state{Nanjing}
%  \country{China}
% }
%\email{ yili_xia@seu.edu.cn}


\begin{abstract}
Along with current multi-scale based detectors, Feature Aggregation and Enhancement (FAE) modules have shown superior performance gains for cutting-edge object detection. However, these hand-crafted FAE modules show inconsistent improvements on face detection, which is mainly due to the significant distribution difference between its training and applying corpus, \textit{i.e.} COCO vs. WIDER Face.
To tackle this problem, we essentially analyse the effect of data distribution, and consequently propose to search an effective FAE architecture, termed AutoFAE by a differentiable architecture search, which outperforms all existing FAE modules in face detection with a considerable margin.
Upon the found AutoFAE and existing backbones, a supernet is further built and trained, which automatically obtains a family of detectors under the different complexity constraints.
Extensive experiments conducted on popular benchmarks, \textit{i.e.} WIDER Face and FDDB, demonstrate the state-of-the-art performance-efficiency trade-off for the proposed automatic and scalable face detector (ASFD) family. In particular, our strong ASFD-D$6$ outperforms the best competitor with AP $96.7/96.2/92.1$ on WIDER Face test, and the lightweight ASFD-D$0$ costs about $3.1$ ms, \textit{i.e.} more than $320$ FPS, on the V100 GPU with VGA-resolution images.
\end{abstract}


\begin{CCSXML}
<ccs2012>
<concept>
<concept_id>10010147.10010178.10010224.10010245.10010250</concept_id>
<concept_desc>Computing methodologies~Object detection</concept_desc>
<concept_significance>500</concept_significance>
</concept>
</ccs2012>
\end{CCSXML}

\ccsdesc[500]{Computing methodologies~Object detection}

\keywords{face detection, neural architecture search, multi-task loss, compound scaling}


\maketitle

\section{Introduction}

\begin{figure}
    \centering
    \includegraphics[width=0.98\linewidth]{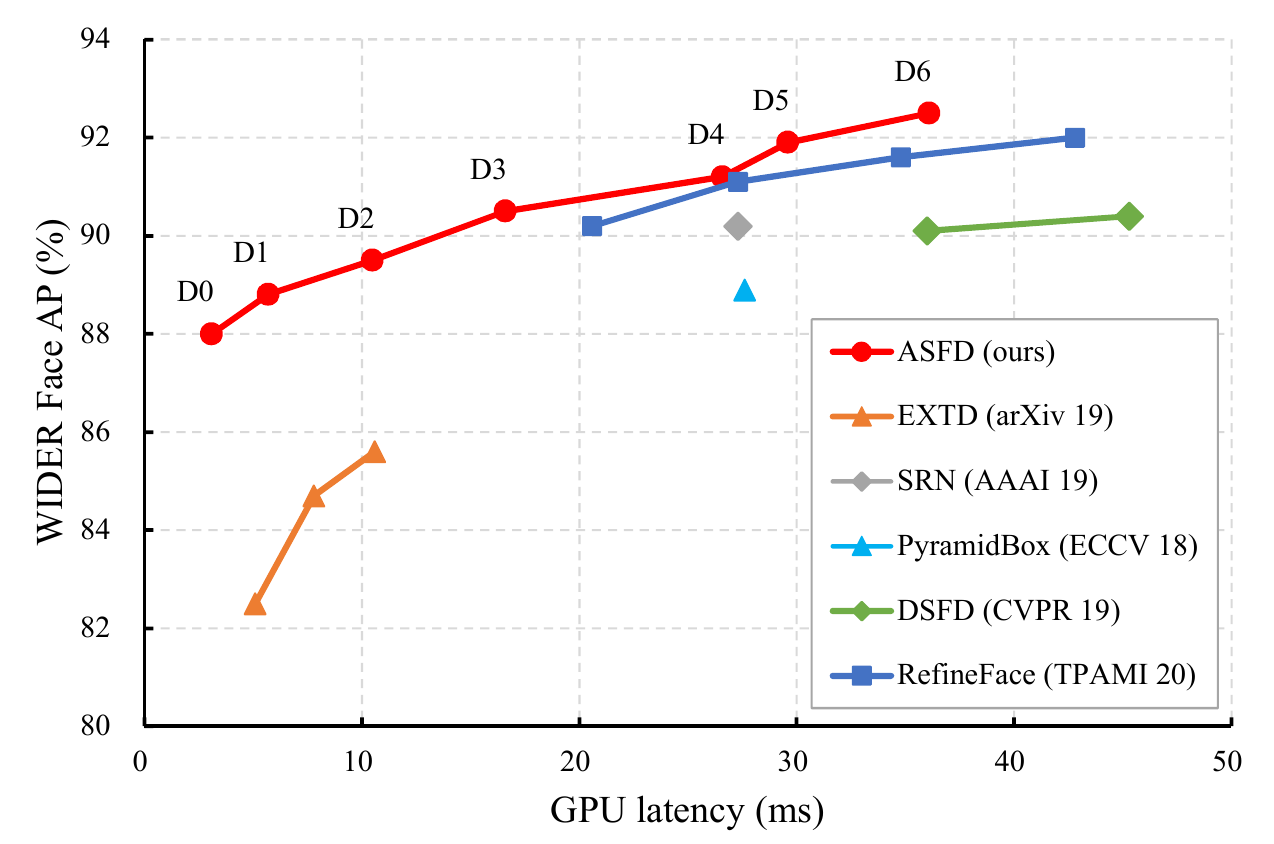}
    \caption{Performance-efficiency trade-off on WIDER Face validation for different face detectors. The proposed ASFD outperforms a range of state-of-the-art methods.}
    \label{fig:trade_off}
\end{figure}

Face detection serves as a fundamental step towards various face-related applications, such as face alignment ~\cite{tai2019towards}, face recognition \cite{huang2020curricularface} and face analysis \cite{pan2018mean}. It aims locate the face region (if any) in a given image, which has been a long standing research topic ranging from \cite{viola2004robust} to deep learning based methods \cite{zhang2017s3fd,chi2019srn}.

% Given an image, the goal of face detection is to determine whether or not there are any faces in it, and if so, to return the locations and shapes of each face. The pioneering work \cite{viola2004robust} adopts the AdaBoost algorithm with hand-crafted features to detect faces, which is undoubtedly not robust especially for those faces attributed to variations in pose, scale, occlusion, expression, \textit{etc} \cite{yang2016wider}. Recently, the face detectors have been significantly improved owing to the strong feature extracting ability of deep Convolution Neural Networks (CNNs) \cite{he2016resnet,simonyan2014vgg}. 
Beyond the scope of face, general object detection has been significantly pushed by the development of deep convolution neural networks \cite{simonyan2014vgg,he2016resnet,ren2015faster,liu2016ssd}. Among one of the representative framework, single-stage anchor-based detector with pyramid features has been thoroughly studied recently \cite{liu2016ssd,lin2017focal} and is dominant for face detection \cite{zhang2017s3fd,chi2019selective,tang2018pyramidbox,li2019dsfd,zhang2020refineface}. In this framework, the regular and dense anchors with different scales and aspect ratios are tiled over all locations of the feature map, and the pyramid features are extracted by the backbone and enhanced by the neck, which is subsequently plugged with both classification and regression branches.

\begin{figure*}[!t]
    \centering
    \subfigure[]{\includegraphics[width=0.32\linewidth]{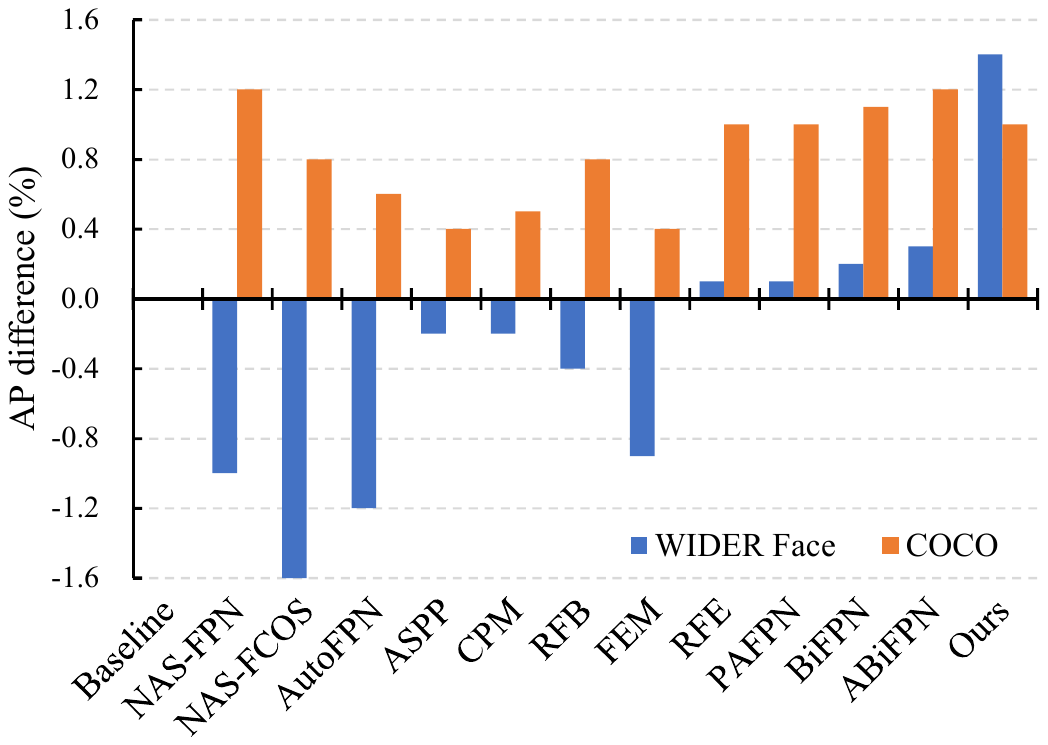}}
    \subfigure[]{\includegraphics[width=0.32\linewidth]{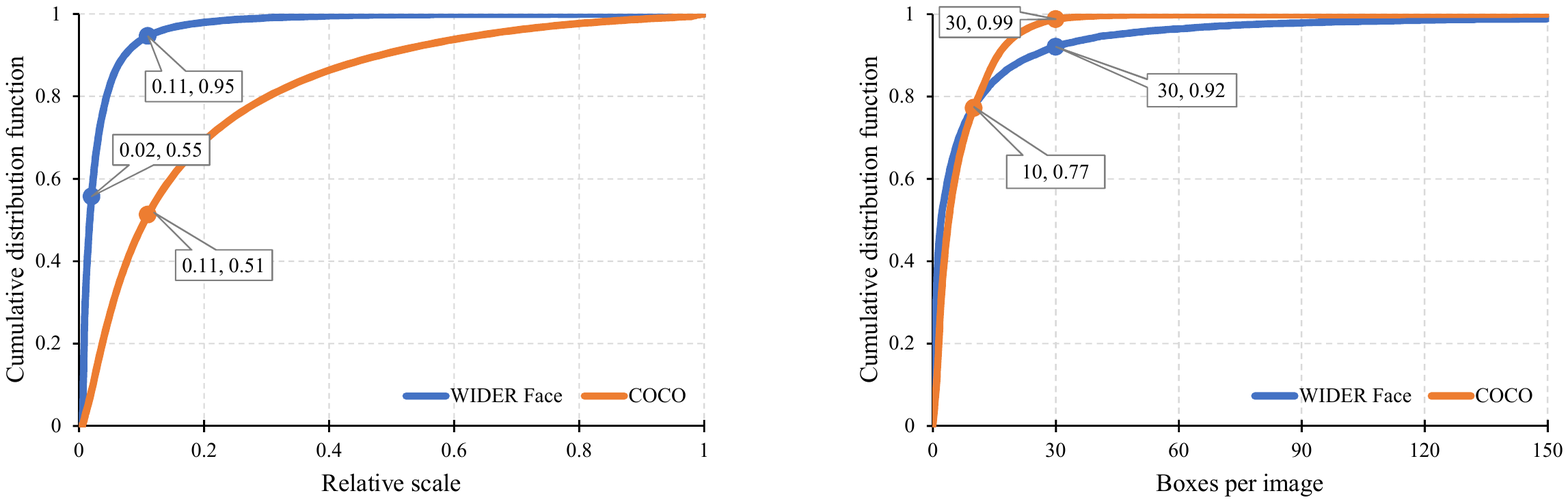}}
    \subfigure[]{\includegraphics[width=0.32\linewidth]{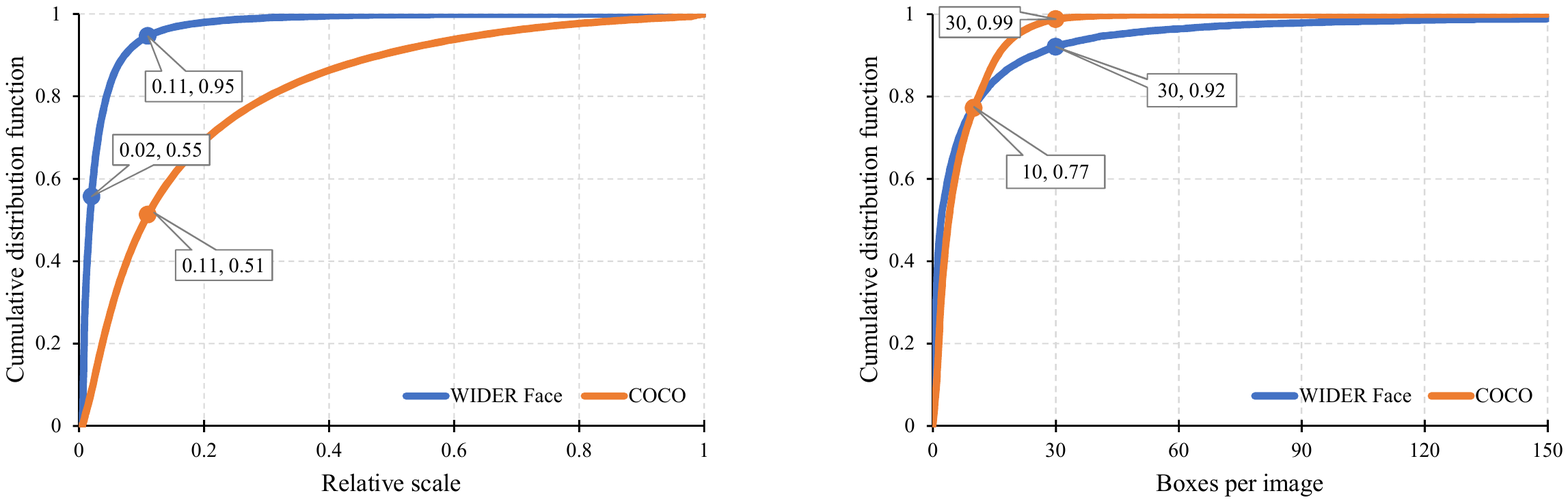}}
    \caption{(a) Comparison of our AutoFAE against other FAE modules on WIDER Face and COCO validation. The performance gaps with the baseline are indicated by blue and orange bars respectively, and RetinaNet is adopted as the baseline.
    (b) Cumulative distribution function (CDF) of the relative scale of bounding boxes. $51\%$ of objects in COCO have a relative scale below $0.11$. For the same scale, the proportion in WIDER Face is $95\%$, while for a similar proportion, $55\%$ of faces in WIDER Face are less than $0.02$.
    (c) CDF of the number of boxes in each image. The distribution of images containing more than $10$ boxes for WIDER Face is long-tailed, \textit{e.g.} $99\%$ of images in COCO have less than $30$ objects, while there are many images in WIDER Face that contain more than $150$ faces.
    }
    \label{fig:gap_coco_widerface}
\end{figure*}

Towards the design of Feature Aggregation and Enhancement (FAE) modules for these methods, Feature Pyramid Network (FPN) and its variants aggregate hierarchical features via the preset pathway, \textit{e.g.} top-down and bottom-up path, to effectively fuse multi-scale features \cite{tang2018pyramidbox,tan2020efficientdet,liu2018pafpn,li2019dsfd,zhang2020acfd}. For another instance, ASPP \cite{chen2017aspp,qiao2020detectors}, RFB \cite{liu2018rfb} and RFE \cite{deng2019retinaface} modules are proposed to enhance the feature representation by adjusting the effective receptive fields. Recently, Neural Architecture Search (NAS) has been also investigated for object detection, which has achieved remarkable performance gains, such as NAS-FPN \cite{ghiasi2019nasfpn}, AutoFPN \cite{xu2019autofpn} and NAS-FCOS \cite{wang2019nasfcos}. However, such a gain is severely not generalized when applying to face detection.

Fig.~\ref{fig:gap_coco_widerface} (a) shows a quantitative investigation of the cutting-edge FAE modules discussed above, in which the significant drops have been shown when they are applied to face domain. Even the automatic learning based method, \textit{a.k.a.} NAS-FCOS \cite{wang2019nasfcos} that performs $1.6$ lower than the baseline. This phenomenon highlights the domain gap between general object and face detection. To explain, we utilize cumulative distribution function to model the corresponding datasets, \textit{e.g.} WIDER Face \cite{yang2016wider} and COCO \cite{lin2014coco} in terms of the relative size of boxes and the number of boxes in each image, as presented in Fig.~\ref{fig:gap_coco_widerface} (b) and (c) respectively. As a result, the relative scale of faces is much smaller than objects in generic object detection, and there are more faces in each image than objects in COCO.
These characteristics also determine the design principles of modern face detectors. For instance, the shallower feature map is adopted to detect the small faces. And more predicted results are retained before and after the non-maximum-suppression for the high recall rate. Since FAE modules designed for generic object detectors are weak when dealing with small-scale and crowded objects, therefore, false positives inevitably exist when they are applied to face domain, resulting in performance degradation.

In this paper, a novel NAS based face detector framework termed Automatic and Scalable Face Detector (ASFD) is introduced, which is designed upon the basis of quantitative observations as above. 
The proposed ASFD is equipped with an effective FAE module, namely AutoFAE, which is discovered in a face-suitable search space, and then automatically scaled up/down to meet different requirements.
In particular, we first analyze why the domain gap between the generic object and face detection would cause such an impact as Fig.~\ref{fig:gap_coco_widerface} (a). The performance degradation in the face domain is caused by the large semantic differences and unreasonable receptive fields for aggregated features.
% of features used for aggregation and unreasonable receptive fields for different levels of features. 
Then, we propose a face-suitable search space that aggregates a feature with similar-scale ones and enriches the feature presentation with different operations for different pyramid levels. And the AutoFAE module is searched by a gradient-NAS method \cite{liu2018darts,xu2019pcdarts}, and can achieve consistent gains on both face detection and generic object detection, as presented in Fig.~\ref{fig:gap_coco_widerface} (a). 
Finally, we build a supernet consisting of the found AutoFAE and a series of backbones, \textit{e.g.} ResNet \cite{he2016resnet}, and automatically obtain the proposed ASFD family to meet different complexity constraints via a one-shot NAS \cite{guo2019spos,chu2019fairnas}.
It is worth noting that the ASFD family achieves the state-of-the-art performance-efficiency trade-off, as presented in Fig.~\ref{fig:trade_off} \cite{yoo2019extd,chi2019srn,tang2018pyramidbox,li2019dsfd,zhang2020refineface}. Especially, the lightweight ASFD-D$0$ can run more than $320$ FPS with VGA-resolution images on a V$100$ GPU, and the strong ASFD-D$6$ obtains the highest AP scores on popular benchmarks, \textit{i.e.} WIDER Face and FDDB. To sum up, this work makes following contributions:
% As presented in Fig.~\ref{fig:trade_off} \cite{yoo2019extd,chi2019srn,tang2018pyramidbox,li2019dsfd,zhang2020refineface}, our automatic and scalable face detector (ASFD) achieves a better balance between performance and efficiency, \textit{i.e.} indicated by GPU latency. In particular, our ASFD-D$0$ can run more than $320$ FPS with VGA-resolution images on a V$100$ GPU, and ASFD-D$6$ obtains the highest AP scores on WIDER Face \cite{yang2016wider} and FDDB \cite{jain2010fddb}, setting a new state-of-the-art face detector. 

\begin{itemize}
    \item We observe an interesting phenomenon that some previous FAE modules perform well in generic object detection but fail in face detection, and conduct extensive experiments to illustrate why this phenomenon occurs.
    \item Based on the observations, we design a face-suitable search space for feature aggregation and enhancement modules, and discover an effective and generalized AutoFAE module via a joint searching method.
    \item Extensive experiments conducted on the popular benchmarks demonstrate the better performance-efficiency trade-off of the proposed ASFD.
\end{itemize}

\section{Related Work}
\subsection{Feature Aggregation and Enhancement.}
In recent years, generic object detection and face detection have been dominated by deep learning based methods. SSD \cite{liu2016ssd} is the first to predict objects using the multi-scale pyramid features, FPN \cite{lin2017fpn} proposes to enrich the feature presentation of multi-scale features by a top-down pathway. Recently, many works are devoted to how to aggregate and enhance multi-scale features effectively. \cite{liu2018pafpn} and \cite{tan2020efficientdet} enhance the entire feature hierarchy by the bottom-up path augmentation. \cite{qiao2020detectors} proposes a novel recursive FPN that incorporates extra feedback connections from FPN into the bottom-up backbone layers. 
% Nowadays, NAS-based methods have demonstrated much success in exploring network architecture. AutoFPN \cite{xu2019autofpn} and NAS-FCOS \cite{wang2019nasfcos} build a fully-connected space to search the reasonable connections and operations among the pyramid levels, and \cite{ghiasi2019nasfpn} achieves this by reinforcement learning (RL). 
Nowadays, NAS-based methods have demonstrated much success in exploring a better architecture for feature fusion and refinement \cite{xu2019autofpn,wang2019nasfcos,ghiasi2019nasfpn}.
Besides, feature enhancement modules are also be widely studied. Inception \cite{szegedy2017inception,szegedy2016rethinking} aims to capture different size of receptive fields via a multi-branch structure. \cite{zhang2020refineface} introduces rectangle receptive fields by a novel enhancement module. \cite{qiao2020detectors,li2019dsfd,liu2018rfb} adopt dilated convolution with different rates to enhance the feature discriminability and robustness. However, as illustrated in Fig.~\ref{fig:gap_coco_widerface}, some of them seem to be ineffective in face detection.

\begin{table}[!t]
    \centering
    \begin{tabular}{c|cccccc}
        \toprule[1pt]
        Pyramid Level & P2 & P3 & P4 & P5 & P6 & P7 \\
        \midrule[0.5pt]
        P2 & $82.9$ & \textcolor{red}{$84.3$} & $85.0$ & $84.7$ & $84.5$ & $82.7$ \\
        P3 & \textcolor{blue}{$83.0$} & $82.9$ & \textcolor{red}{$85.1$} & $84.8$ & $84.5$ & $83.2$ \\
        P4 & $83.0$ & \textcolor{blue}{$83.2$} & $82.9$ & \textcolor{red}{$84.5$} & $83.8$ & $83.3$\\
        P5 & $82.7$ & $83.0$ & \textcolor{blue}{$83.0$} & $82.9$ & \textcolor{red}{$83.7$} & $83.2$ \\
        P6 & $82.8$ & $83.0$ & $82.9$ & \textcolor{blue}{$82.9$} & $82.9$ & \textcolor{red}{$83.1$} \\
        P7 & $82.7$ & $82.8$ & $83.3$ & $83.0$ & \textcolor{blue}{$83.0$} & $82.9$ \\
        \bottomrule[1pt]
    \end{tabular}
    %\vspace{-1mm}
    \caption{ Performance of FPN on Hard subset of WIDER Face validation while a pyramid level (indicated by the row) is aggregated by a specific level (indicated by the column).}
    \label{tab:pyramid_fa}
\end{table}

\subsection{Neural Architecture Search.}
NAS first uses reinforcement learning to search for hyper-parameters in the structure or used in the training process \cite{zoph2016neural,zoph2018learning,tan2019efficientnet}. Recent researches focus on the automatic search of network architecture. Based on the idea of weight sharing, some works try to build the final structure by stacking a searched cell several times \cite{liu2018darts,xu2019pcdarts}, and other methods \cite{guo2019spos,chu2019fairnas,cai2019once,liu2020bfbox,chen2019detnas} decouple the training and searching process and directly train a supernet by randomly sampling a single-path network at each time.
As for their applications on object detection to fuse the multi-scale features, NAS-FPN \cite{ghiasi2019nasfpn} searches the irregular connections among pyramid layers with an RNN controller for aggregating the multi-scale features. AutoFPN \cite{xu2019autofpn} and NAS-FCOS \cite{wang2019nasfcos} discover the aggregation modules within a fully-connected search space densely connecting any two layers, in which features from some layers that damage the aggregated feature may be introduced causing accuracy degradation.
BFBox \cite{liu2020bfbox} is the first attempt of NAS on face detection and proposes a face-suitable search space. Although the novel backbone and neck networks are discovered among the search space, its performance is still worse than the state-of-the-art face detectors.

% NAS can be roughly divided into three categories, RL-based \cite{zoph2016neural,zoph2018learning}, evolutionary algorithm (EA)-based \cite{guo2019spos,chu2019fairnas} and gradient-based \cite{liu2018darts,xu2019pcdarts}. Owing to the lower computation requirements of EA- and gradient-based methods, they are widely used in object detection for automatically searching some components, such as backbone, neck and head networks. \cite{chen2019detnas} introduce a lightweight backbone searching approach for object detection, and \cite{liu2020bfbox} propose to enhance the features from shallow layers of face detection. \cite{ghiasi2019nasfpn,xu2019autofpn,wang2019nasfcos} aim to discover a better feature fusion module for generic object detection.

% In this paper, a face-suitable search space is proposed for discovering the better feature aggregation and enhancement module. Particularly, the cross-scale connections are searched along the preset pathway rather than in a violent fully-connected manner, and different FE modules are discovered to suit the different scales of pyramid features. By relaxing the architectures into several parameters, they are updated by a gradient-based NAS approach using bi-level optimization.

In this work, the sparse cross-scale connections of FA module are searched based on a face-suitable search space rather than in a violent fully-connected manner. And various FE modules with different operations and topologies are discovered for different pyramid levels.

\section{Problem Analysis}\label{sec:problem}
Fig.~\ref{fig:gap_coco_widerface} (a) is sufficient to illustrate the inconsistency between general object detection and face detection. In order to further analyze the reason why this phenomenon occurs, the effects of feature aggregation and enhancement modules are discussed respectively in this section.

\subsection{Feature Aggregation (FA).}
Firstly, extensive experiments are conducted to explore the relationship between performance and cross-connection of FA modules. These experiments are to add a FA module (for simplicity, FPN \cite{lin2017fpn}) in turn between any two pyramid features and aggregate one feature with another one after resizing to the same shape. Table~\ref{tab:pyramid_fa} shows the results on the diagonal indicating RetinaNet without FPN, and upper triangle and lower triangle representing top-down and bottom-up paths respectively, especially, red and blue fonts indicate top-down and bottom-up paths in FPN and PAFPN. it is clear to conclude that aggregating multi-scale features through top-down paths is superior to the bottom-up ones, especially these two layers used are close to each other. As the distance increasing, some connections even cause performance degradation, \textit{e.g.} AP$_{.50}$ drops $0.2$ when P$2$ is aggregated by P$7$.
% As a counterpart, even some connections may cause performance degradation as the distance increases.
Therefore, small faces only occur in the shallow features cannot be enhanced by semantic-rich features with large scale difference.
It reveals that NAS-FPN, AutoFPN and NAS-FCOS are sub-optimal to fuse features through a fully-connected or irregular connection in the face domain. 
% In addition, aggregating multi-scale features through the adjacent layers \cite{lin2017fpn,liu2018pafpn,tan2020efficientdet} is also sub-optimal especially for the shallow layers, although it has improved the performance with a significant margin. For instance, an obvious performance advantage exists when P$2$ is aggregated using P$4$ rather than the adjacent P$3$. 

\begin{table}[!t]
    \centering
    \begin{tabular}{l|cccccc}
        \toprule[1pt]
        Module & P2 & P3 & P4 & P5 & P6 & P7 \\
        \midrule[0.5pt]
        ASPP & $86.5$ & $86.8$ & $86.9$ & $87.2$ & $\mathbf{87.5}$ & $87.1$ \\
        CPM & $86.8$ & $86.9$ & $87.0$ & $87.1$ & $\mathbf{87.4}$ & $87.2$ \\
        RFB & $86.8$ & $86.9$ & $87.0$ & $87.3$ & $\mathbf{87.4}$ & $87.2$ \\
        RFE & $87.4$ & $\mathbf{87.6}$ & $87.5$ & $87.4$ & $87.2$ & $87.1$ \\
        \bottomrule[1pt]
    \end{tabular}
    %\vspace{-1mm}
    \caption{Performance of different feature enhancement modules when operated on different pyramid levels.}
    \label{tab:pyramid_fe}
\end{table}

\subsection{Feature Enhancement (FE).}
Similar experiments are conducted to demonstrate the effects of different feature enhancements modules. As shown in Table~\ref{tab:pyramid_fe}, there are significant performance differences when a FE module is applied to the different pyramid layers. 
% even this gap can reach $1$ point when ASPP \cite{chen2017aspp,qiao2020detectors} is operated on P$2$ and P$6$ respectively.
In general, ASPP \cite{chen2017aspp,qiao2020detectors}, CPM \cite{li2019dsfd,tang2018pyramidbox} and RFB \cite{liu2018rfb} employ dilated convolution with different rates to enlarge the receptive fields, they can obtain the consistent performance when applied to different pyramid layers. Particularly, they would damage the shallow features especially the first two layers, and cause severe performance degradation.
RFE \cite{zhang2020refineface} aims to enrich the features by introducing the rectangle receptive fields and performs well on all pyramid layers especially the shallow layers. 
A meaningful conclusion can be drawn that the shallower features seem to prefer a more diverse receptive field while the deeper layers favor a larger one. 
This is mainly because detecting faces with occlusion or extreme-pose that appear in the shallow layers expect more robust features, and the large faces require features with large receptive fields to locate accurately.

In summary, reasons for the aforementioned problem are: (1) \textit{The unreasonable connection in FA modules would cause performance degradation}, (2) \textit{Features from different layers should be enhanced by different operations}.
% These discussions encourage us to design a suitable search space for face detection instead of simply inheriting settings from image classification, it is detailed described in the next section.

\section{Methodology}
The framework of our ASFD is based on the simple and effective RetinaNet \cite{lin2017focal}, which contains three main components: the backbone for extracting pyramid features, the neck for fusing and enhancing the features, and the head for regression and classification. Our goal is to discover a \textit{better neck architecture} for RetinaNet and scale the ASFD to satisfy different complexity requirements automatically. 

\subsection{Search Space of AutoFA and AutoFE}
\subsubsection{AutoFA}
% \textbf{To address the problem}
% Ablative experiments in Table~\ref{tab:pyramid_fa} have shown the effects of different connections of FPN, from which we have drawn two clear conclusions. Firstly, shallow features are rich in texture information, as the network deepens, semantic information conducive to prediction is gradually enriched. Thus, aggregating the shallow feature with those semantically rich features brings performance improvements with a large margin, \textit{i.e.} merging along a top-down pathway. Secondly, it is not that the richer the semantic information used for fusion, the better performance can be obtained. 
% In general, it is better to aggregate features with similar but not necessarily adjacent scales, performance would drop sharply as the scale difference becomes larger.
In order to address the limitation of previous NAS-based FPN \cite{wang2019nasfcos, xu2019autofpn,ghiasi2019nasfpn} when applied on the face domain, the above analysis motivates us to design a module that aggregates a feature by the similar-scale features instead of directly using those with large differences in scale.

\begin{figure}[!t]
    \centering
    \includegraphics[width=0.95\linewidth]{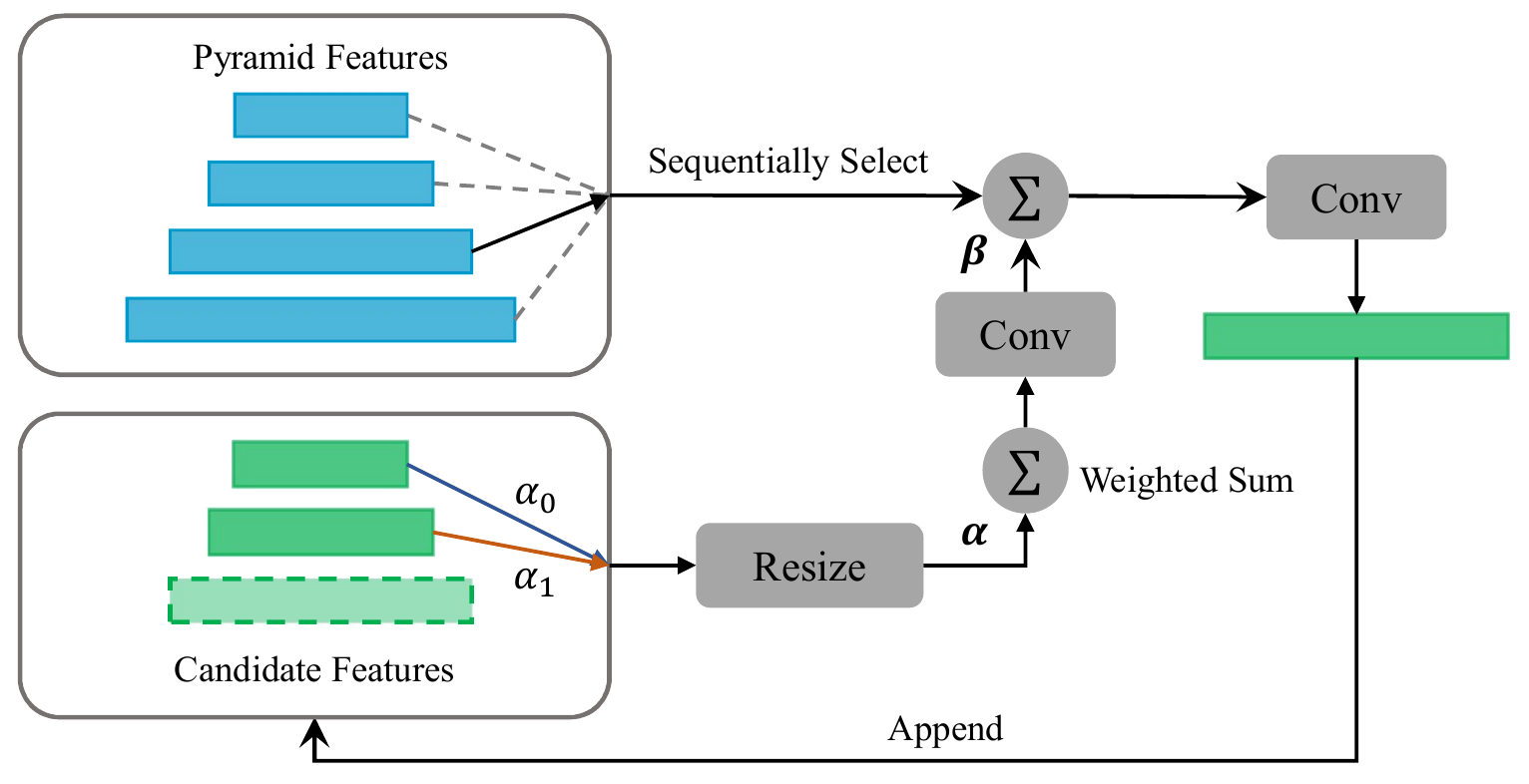}
    \caption{Illustration of the basic cell of AutoFA. $\sum$ means the sum weighted by a factor. $\boldsymbol{\alpha}$ indicates the probability to choose a candidate feature, and $\boldsymbol{\beta}$ is the score for weighting the importance of different features.}
    \label{fig:fpn_uint}
\end{figure}

To this end, we propose a fundamental building cell of AutoFA for aggregating the pyramid features sequentially, shown in Fig.~\ref{fig:fpn_uint}. Initially, the cell contains a pyramid feature pool with a specific one activated, and a candidate feature pool with aggregated features of previous steps. During the searching phase, the specific pyramid feature is selected sequentially, candidate features are chosen with the corresponding probability $\boldsymbol{\alpha}$. Firstly, these candidate features are aggregated together after resizing to the same shape and weighting by $\boldsymbol{\alpha}$; then, it is fused with the pyramid feature and a convolution layer is performed to obtain the corresponding aggregated feature. At last, it is appended to the candidate feature pool for the later feature fusion.
Assume that a pyramid feature and the corresponding aggregated feature are $\mathbf{F}_i$ and $\mathbf{C}_i$, the basic cell can be formulated as,

\begin{equation}
    \mathbf{C}_i = f_{post}\left(\beta_0 \mathbf{F}_i + \beta_1 f_{pre}\left(\sum\nolimits_{j<i}\alpha_j f_{re}(\mathbf{C}_j)\right)\right),
\end{equation}

in which $\sum_{j<i}\alpha_j\!=\!1$, $f_{post}(\cdot)$ and $f_{pre}(\cdot)$ are two convolution operations for feature aggregation, and $f_{re}(\cdot)$ is for resizing the feature to the same size, \textit{i.e.} bilinear interpolation for upsampling and maxpooling with stride $2$ for downsampling. Once the searching process is done, the final discrete structure can be obtained according to the probability score $\boldsymbol{\alpha}$ and importance score $\boldsymbol{\beta}$. For a given pyramid feature, it is aggregated by candidate features with probability $\alpha_i\geq0.5$, and $\boldsymbol{\beta}$ is retained as the initial value for weighting the pyramid feature and candidate feature. In this approach, aggregated features of the discrete cell can be denoted as,
\begin{equation}
\begin{split}
    \mathbf{T}_i &= f_{pre}\left(\sum\nolimits_{j<i}[\alpha_j\geq0.5]\cdot f_{re}(\mathbf{C}_j)\right), \\
    \mathbf{C}_i &= f_{post}\left(\beta_0\cdot\mathbf{F}_i + \beta_1\cdot\mathbf{T}_i\right),
\end{split}
\end{equation}
where $[\!~\cdot~\!]$ equals $1$ if the inner expression is true.

Similar to PAFPN \cite{liu2018pafpn} and BiFPN \cite{tan2020efficientdet}, our AutoFA aggregates the pyramid features along a top-down path and a bottom-up path, each of them is comprised of several basic building cells. For the top-down path, pyramid features are selected in the order of decreasing resolution for aggregation, \textit{i.e.} from P$7$ with stride $128$ to P$2$ with stride $4$, same as Fig.~\ref{fig:fpn_uint}. As the counterpart, the aggregation along bottom-up path is in the reversed order.

\subsubsection{AutoFE.}
% Extensive experiments in Table~\ref{tab:pyramid_fe} have revealed the incompatibility of these modules for some level of features and drawn an important conclusion that features from different pyramid layers expect different types of receptive fields and the specific enhancement modules should be designed for them separately.
\begin{figure}[!t]
    \centering
    \includegraphics[width=0.99\linewidth]{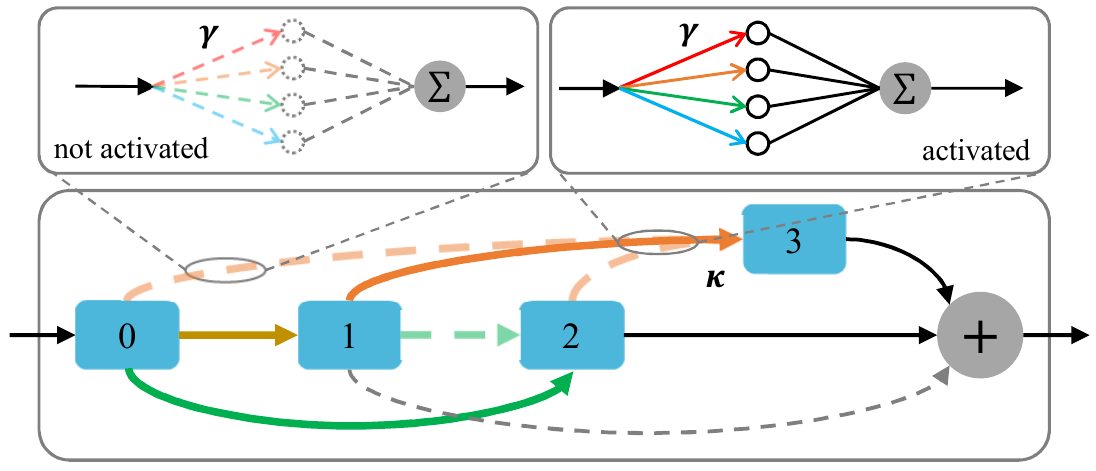}
    \caption{Illustration of the basic structure of AutoFE with $4$ nodes. The bold colored arrows have two states: not activated and activated, in which not activated arrows mean disconnecting, $\boldsymbol{\kappa}$ indicates the probability. Thin colored arrows indicate different operations with probability $\boldsymbol{\gamma}$.}
    \label{fig:fe_uint}
\end{figure}

The incompatibility of those FE modules for some pyramid levels has been revealed and an important conclusion has been drawn in the aforementioned analysis.
To discover the suitable enhancement module for each pyramid layer, we propose a basic cell for our AutoFE that includes several intermediate features transformed by the candidate operations, which include \{$1\times 1$ conv, $1\times 3$ conv, $3\times 1$ conv, $3\times 3$ conv, $1\times 5$ conv, $5\times 1$ conv, $5\times 5$ conv \}. As presented in Fig.~\ref{fig:fe_uint}, the basic cell is conducted as a directed acyclic graph with several nodes, where node $0$ is input and others are intermediate features. 
Each node $i$ is connected to the previous node $j\!<\!i$ with two status indicated by $\kappa_{ji}$, \textit{i.e.} activated if and only if $\kappa_{ji}$ is maximum among $\kappa_{\ast i}$, otherwise not activated.
% the connection is activated if and only if $\kappa_{ji}$ is maximum among $\kappa_{\ast i}$. 
In this way, the previous feature is transformed by the different operations; otherwise, it is not activated. Assume that the feature of $i$th node is $\mathbf{F}_i$, it can be formulated as follow,
\begin{equation}\label{eq:illu_fe_unit}
    \mathbf{F}_i = \sum\nolimits_{j<i} [j\!=\!\mathop{\arg\max}_j \kappa_{ji}]\cdot f_{op}(\mathbf{F}_j, \gamma_{ji}),
\end{equation}
where $f_{op}(\cdot)$ is the sum weighted by $\gamma_{ji}$ when processed by the activated operations. Different from \cite{liu2018darts,xu2019pcdarts}, the output of the cell is the sum of features of all leaf nodes, \textit{i.e.} the intermediate features who are not input to the other nodes, given by
\begin{equation}
    \mathbf{F}_{out} = \sum\nolimits_i \left[\sum\nolimits_{k>i} [i\!=\!\mathop{\arg\max}_i \kappa_{ik}] = 0\right] \cdot \mathbf{F}_i.
\end{equation}
In particular, the commonly used convolutions with different kernel shapes and dilation rates are adopted for $f_{op}(\cdot)$.

However, during the search, Eq.~\ref{eq:illu_fe_unit} cannot be optimized because it is equivalent to discrete sampling, which is not differentiable. To allow back-propagation, we use the Gumbel-Max method \cite{dong2019gdas} to re-formulate Eq.~\ref{eq:illu_fe_unit} in an efficient way that samples a discrete probability as follow,
% \begin{small}
\begin{equation}\label{eq:gumbel}
    \begin{split}
    \mathbf{F}_i &= \sum\nolimits_{j<i} h_{ji}\cdot f_{op}(\mathbf{F}_j, \gamma_{ji}), \\
    \text{s.t.} \quad h_{ji} &= \text{onehot}(\mathop{\arg\max}_j(\kappa_{ji} + o_{ji})),
    \end{split}
\end{equation}
% \end{small}
where $o_{ji}$ is the \textit{i.i.d.} sample drawn from Gumbel$(0,1)$ \cite{dong2019gdas}. Then, softmax function is used to relax the argmax function so as to make Eq.~\ref{eq:gumbel} being differentiable, in which $\tilde{h}_{ji}$ is for approximating $h_{ji}$, denoted by,
% \begin{small}
\begin{equation}\label{eq:softmax}
    \tilde{h}_{ji} = \frac{\exp{(\kappa_{ji} + o_{ji}/\tau)}}{\sum\nolimits_{j'<i}\exp{(\kappa_{j'i} + o_{j'i}/\tau)}},
\end{equation}
% \end{small}
where $\tau$ is the softmax temperature. In this way, argmax is used in the forward pass to achieve discrete sampling of connections between two nodes, but softmax in Eq.~\ref{eq:softmax} is adopted during backward pass to allow gradient back-propagation.

Finally, the discrete architecture of AutoFE is obtained by retaining the connections and operations among intermediate features according to the maximum of $\boldsymbol{\kappa}$ and $\boldsymbol{\gamma}$.

% The modern face detectors \cite{najibi2017ssh, deng2019retinaface,tang2018pyramidbox,li2019dsfd} usually enrich the feature presentation by an extra module due to the lack of semantics of small-scale faces, in which convolution operations with different dilated rates and different kernel shape are widely used to make the receptive fields enlarge and more diverse respectively.
\subsection{Search Strategy of AutoFA and AutoFE}
We have transformed the discrete network structure into several architecture parameters through the design of face-suitable search space for AutoFA and AutoFE. In detail, $\boldsymbol{\alpha}$ is adopted to make the decision on choosing candidate features for a pyramid feature, $\boldsymbol{\beta}$ is used to balance the importance of pyramid and candidate features. For AutoFE, $\boldsymbol{\kappa}$ is employed for selecting the connection of intermediate nodes, and $\boldsymbol{\gamma}$ indicates the probability of different operations. Similar to \cite{liu2018darts,chen2019pdarts,xu2019pcdarts}, we utilize the bi-level optimization method to alternately optimize the network parameters, \textit{e.g.} parameters of convolution layers, and architecture parameters in an end-to-end manner.

\begin{figure}[!t]
    \centering
    \includegraphics[width=0.9\linewidth]{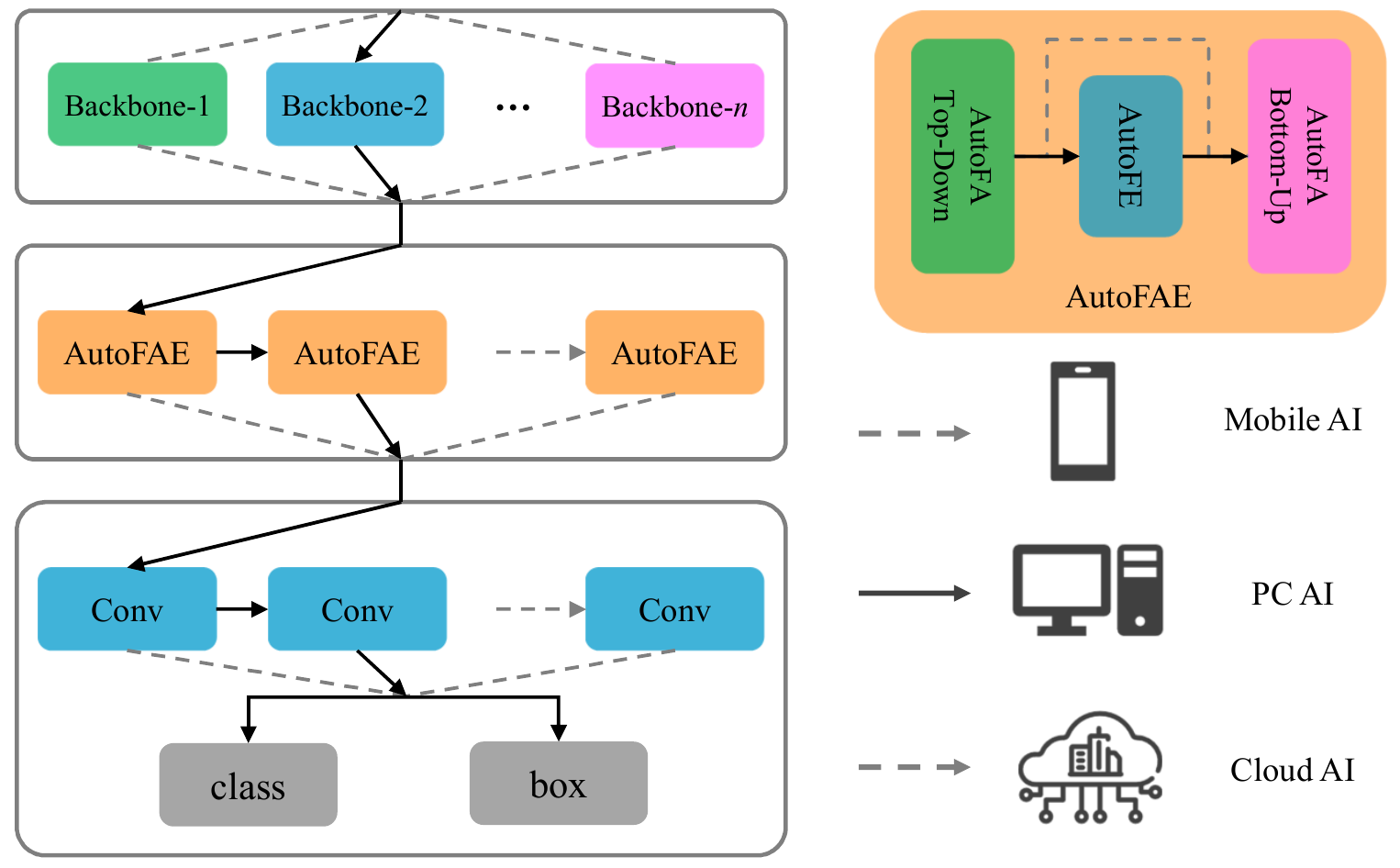}
    \caption{The architecture of the supernet to automatically obtain a detector for different AI systems.}
    \label{fig:arch_supernet}
\end{figure}

\subsection{Auto Model Scaling}
% Recently, one-shot NAS \cite{guo2019spos,chu2019fairnas} is proposed with the idea that decoupling the supernet training and architecture searching in two sequential steps, which allows training the supernet once, then searching for multiple architectures. 
% Inspired by the success of one-shot NAS \cite{guo2019spos,chu2019fairnas}, we design a more flexible search space to achieve model scaling automatically. In particular, we build a supernet that forms by stacking multiple backbone networks in parallel and stacking the discovered AutoFAE and prediction convolution in cascade as candidates, as shown in Fig.~\ref{fig:arch_supernet}.

We automatically obtain the ASFD family with different complexities on the basis of a supernet, as shown in Fig.~\ref{fig:arch_supernet}, which is comprised of the backbones in parallel, the stacked AutoFAE modules, and the stacked convolutions for prediction head. Our method aims to search for a better composition to meet different complexity requirements, \textit{i.e.} which backbone to pick, how many AutoFAE modules and convolutions in head to stack, whether to skip AutoFE for each AutoFAE, and what the number of feature channels.

% \vspace{-4mm}
\subsubsection{Training.}
Based on the idea of weight sharing, the supernet is trained by alternately training a single-path network through uniformly sampling \cite{guo2019spos,chu2019fairnas}.
% Similar to \cite{guo2019spos} and \cite{chu2019fairnas}, our supernet is trained in a simple approach. Specifically, based on the opinion of weight sharing \cite{guo2019spos}, a single-path is sampled uniformly to update at each iteration, 
For instance, as presented in Fig.~\ref{fig:arch_supernet}, the single-path is composed of backbone-$2$, and AutoFAE and prediction convolutions, which are both stacked two layers. 
Furthermore, a scalable method is proposed for training the supernet compatible with different feature channels. The supernet is optimized with the maximal feature channels during a long warm-up period until it tends to converge. Then, the candidate feature channels are gradually added to be sampled in the descending order, in which the corresponding tensors are sliced out along each dimension to fit the calculations.

\subsubsection{Searching.}
The searching phase is based on the genetic algorithm \cite{guo2019spos,chen2019detnas,chu2019fairnas} and directly takes inference latency into fitness. At first, populations are randomly initialized with genes encoded by the $5$ degrees of freedom of the supernet, which would be removed if against the constraints. After the initialization, they are evaluated on a mini validation set to obtain the fitness. At each iteration, only the top-$k$ populations with better finesses are retained to generate the next generation by mutation and crossover. By repeating this procedure several times, we can discover a single-path network with the best fitness.

\section{Experiments}
\subsection{Experimental Setup}
\subsubsection{Baseline.} If not specified, RetinaNet \cite{lin2017focal} with FPN is utilized as the baseline of the face detector. Compared to the original generic object detection application, it has the following differences: (1) 6 levels of pyramid features are used for predicting with anchor scales $\{4,8,16,32,64,128\}$ and aspect ratio $1\!\!:\!\!1.5$. (2) The IoU threshold for anchor matching is changed to $0.4$ and the ignore-zone is not implemented. (3) Top-2000 predictions with confidence higher than $0.05$ are processed by non-maximum suppression with a threshold $0.4$ to produce at most $750$ final detections. 
The results are reported using AP$_{.50}$ measured with a constant IoU threshold $0.5$, as well as AP averaged under IoU thresholds from $0.5$ to $0.95$ with step $0.05$ to demonstrate the performance at high IoU.

\subsubsection{Train Details.} We use the ImageNet-pretrained models to initialize the backbone parameters, and `kaiming' method for others. SGD algorithm is employed to optimize the network parameters with momentum $0.9$, weight decay $5\!\times\!10^{-4}$ and initial learning rate $0.01$ per $32$ images. For ablative studies, the learning rate is multiplied by factor $0.1$ at $30$, $40$ epochs and ended at $50$ epochs.
For the main results, it is divided by $10$ at $60$, $100$ epochs and ended at $120$ epochs.

\subsubsection{Search Details.} The training set of WIDER Face is divided into two mini training and a validating subsets, with a ratio of $9\!:\!9\!:\!2$, they are used for updating network and architecture parameters, and evaluating the searched modules respectively.
Adam algorithm with learning rate $0.01$ is adopted for optimizing the architecture parameters, which are frozen at the first $50$ epochs and updated during $50\!\sim\!100$ epochs, and other settings are same as training details. To determine the final AutoFAE module, we run the searching algorithm $3$ times with different random seeds and pick the best one based on its performance on the mini validation. All training and searching experiments are conducted on 8 V100 GPUs. The AutoFA and AutoFE can be searched within 3 to 4 hours. The commonly used supernet takes about 12 hours for training, and the ASFD families could be sampled within 1.5 to 4 hours.

\subsection{Ablation Study}

\begin{table}[!t]
    \centering
    \begin{tabular}{l|ccc|ccc}
        \toprule[1pt]
        \multirow{2}{*}{Module} & \multicolumn{3}{c|}{AP$_{.50}$} & \multicolumn{3}{c}{AP} \\
        \cline{2-7}
        & Easy & Medium & Hard & Easy & Medium & Hard \\
        \midrule[0.5pt]
        Baseline & $95.1$ & $94.0$ & $87.2$ & $61.9$ & $59.2$ & $46.5$ \\
        NAS-FPN & $95.1$ & $93.9$ & $86.2$ & $61.8$ & $59.0$ & $45.7$ \\
        NAS-FCOS & $94.7$ & $93.1$ & $85.6$ & $61.5$ & $58.1$ & $45.1$ \\
        AutoFPN & $94.6$ & $93.4$ & $86.0$ & $61.4$ & $58.5$ & $45.6$ \\
        PAFPN & $95.3$ & $94.1$ & $87.3$ & $62.1$ & $59.4$ & $46.8$ \\
        BiFPN & $95.5$ & $94.4$ & $87.4$ & $62.3$ & $59.6$ & $46.9$ \\
        ABiFPN & $95.3$ & $94.5$ & $87.5$ & $62.2$ & $59.7$ & $47.0$ \\
        FEM-FPN & $95.2$ & $94.0$ & $86.7$ & $62.1$ & $59.4$ & $46.5$ \\
        DARTS & $95.1$ & $93.5$ & $86.5$ & $61.8$ & $58.6$ & $45.6$ \\
        PC-DARTS & $95.0$ & $93.7$ & $86.6$ & $61.8$ & $58.9$ & $46.0$ \\
        \bottomrule[0.5pt]
        AutoFA & $95.4$ & $94.4$ & $\mathbf{87.8}$ & $\mathbf{62.8}$ & $\mathbf{60.2}$ & $\mathbf{47.4}$\\
        \bottomrule[1pt]
    \end{tabular}
    \caption{Comparison with state-of-the-art feature aggregation modules on WIDER Face validation.}
    \label{tab:autofa}
\end{table}

\subsubsection{Effect of Search Space for AutoFA and AutoFE}
To demonstrate the effectiveness of our proposed face-suitable search space for feature aggregation and enhancement modules, the AutoFA and AutoFE modules are discovered and compared to the state-of-the-art modules respectively.

At the first stage, the AutoFA module is searched through a RetinaNet that replaces the FPN with several basic aggregation modules. 
As shown in Table~\ref{tab:autofa}, simulations are conducted by comparing to the commonly used FA modules, in which DARTS \cite{liu2018darts} and PC-DARTS \cite{xu2019pcdarts} illustrate the results based on a fully-connected search space \cite{wang2019nasfcos}. Our AutoFA manages to address the limitations of previous NAS-based methods and outperforms them with a large margin, which is more than $1.0$ points on all three subsets indicated by AP. 
Besides, it is also significantly better than the hand-crafted ones composed of top-down and bottom-up paths, \textit{i.e.} PAFPN \cite{liu2018pafpn}, BiFPN \cite{tan2020efficientdet}, and ABiFPN \cite{zhang2020acfd}, demonstrating the superiority of connections between the multi-scale features of AutoFA.
Such the large improvement is mainly from predictions with the high IoU, which shows that the features of different scales are fully aggregated and it is helpful for more distinguishable classification and more accurate location.

\begin{table}[!t]
    \centering
    \begin{tabular}{l|ccc|ccc}
        \toprule[1pt]
        \multirow{2}{*}{Module} & \multicolumn{3}{c|}{AP$_{.50}$} & \multicolumn{3}{c}{AP} \\
        \cline{2-7}
        & Easy & Medium & Hard & Easy & Medium & Hard \\
        \midrule[0.5pt]
        Baseline & $94.7$ & $92.8$ & $82.9$ & $61.7$ & $58.4$ & $45.0$ \\
        ASPP & $94.8$ & $93.0$ & $83.4$ & $62.1$ & $58.8$ & $45.3$ \\
        RFB & $94.5$ & $92.7$ & $83.0$ & $61.5$ & $58.5$ & $45.2$ \\
        CPM & $94.6$ & $92.8$ & $83.0$ & $61.5$ & $58.4$ & $45.2$ \\
        FEM-CPM & $94.5$ & $92.9$ & $83.3$ & $61.9$ & $58.7$ & $45.5$ \\
        RFE & $94.5$ & $92.8$ & $83.2$ & $61.8$ & $58.7$ & $45.4$ \\
        DARTS & $94.7$ & $92.9$ & $83.0$ & $61.6$ & $58.6$ & $45.1$ \\
        PC-DARTS & $94.6$ & $93.0$ & $83.0$ & $61.8$ & $58.6$ & $45.2$ \\
        \bottomrule[0.5pt]
        AutoFE & $\mathbf{95.2}$ & $\mathbf{93.2}$ & $\mathbf{83.5}$ & $\mathbf{62.1}$ & $\mathbf{59.0}$ & $\mathbf{45.8}$ \\
        \bottomrule[1pt]
    \end{tabular}
    \caption{Comparison with state-of-the-art feature enhancement modules on WIDER Face validation.}
    \label{tab:autofe}
\end{table}

Then, RetinaNet without FPN is adopted as the baseline to better highlight the effectiveness of FE modules. 
Different FE modules are placed between the backbone and detection head to refine the multi-scale features, as shown in Table~\ref{tab:autofe}. In particular, DARTS and PC-DARTS discover FE modules by following their original settings in image classification. However, they only improve the baseline by minor advantages.
With the specified face-suitable search space, the found AutoFE improves the baseline by $0.5/0.4/0.6$ points of AP$_{.50}$ and $0.4/0.6/0.8$ points of AP, far exceeding the other state-of-the-art modules and demonstrating the superiority of our face-suitable search space.

\subsubsection{Effect of Joint Searching AutoFAE}
The AutoFAE module is composed of AutoFA and AutoFE two modules, which can be obtained by cascading the discovered AutoFA and AutoFE modules or jointly searching in an end-to-end manner. 
As presented in Table~\ref{tab:joint_search}, only a minor improvement is achieved by cascading the discovered AutoFA and AutoFE directly. And AutoFAE found by the joint searching way can further improve AP$_{.50}$ and AP by clear margins, demonstrating the state-of-the-art performance of proposed AutoFAE.

% compared to cascade AutoFA and AutoFE together, AutoFAE found by the joint searching way can further improve the performance by a clear margin, which demonstrates its advantage.
% Table~\ref{tab:joint_search} shows the effectiveness of different searching methods of AutoFAE, in which joint searching  can further improve the baseline with considerable gaps compared to the cascade one and demonstrates the advantage of our method. 

\begin{table}[!t]
    \centering
    \resizebox{0.99\linewidth}{!}{
    \begin{tabular}{l|ccc|ccc}
        \toprule[1pt]
        \multirow{2}{*}{Method} & \multicolumn{3}{c|}{AP$_{.50}$} & \multicolumn{3}{c}{AP} \\
        \cline{2-7}
        & Easy & Medium & Hard & Easy & Medium & Hard \\
        \midrule[0.5pt]
        Baseline & $95.1$ & $94.0$ & $87.2$ & $61.9$ & $59.2$ & $46.5$ \\
        % \midrule[0.5pt]
        % DARTS & $95.4$ & $94.2$ & $87.4$ & $62.0$ & $59.4$ & $46.8$ \\
        % PC-DARTS & $95.3$ & $94.4$ & $87.5$ & $62.0$ & $59.5$ & $47.0$ \\
        % \midrule[0.5pt]
        AutoFA+AutoFE & $95.4$ & $94.5$ & $87.9$ & $62.8$ & $60.2$ & $47.5$ \\
        Joint Search & $\mathbf{95.7}$ & $\mathbf{95.0}$ & $\mathbf{88.6}$ & $\mathbf{62.9}$ & $\mathbf{60.5}$ & $\mathbf{47.8}$ \\
        \bottomrule[1pt]
    \end{tabular}}
    \caption{The effect of searching method for the AutoFAE.}
    \label{tab:joint_search}
\end{table}

\subsubsection{Effect of Different Positions of AutoFE}
Review that our AutoFAE is built upon the top-down and bottom-up paths. Therefore, we have three ways to build the final AutoFAE module. In detail, the AutoFE module can be plugged before and after the AutoFA, as well as between the top-down and bottom-up paths. In this way, three modules are obtained by utilizing the joint searching method. As shown in Table~\ref{tab:different_position}, we observe that the best performance is achieved when AutoFE is in the middle position. This is mainly because similar presentation is generated after the top-down aggregation. Placing AutoFE before the bottom-up path can further enhance these features to carry different context information.
% This is mainly because the similar contextures are generated after the top-down aggregations, AutoFE module can enrich the multi-scale features with different presentations when it is placed before the bottom-up aggregating path. 

\begin{table}[!t]
    \centering
    \begin{tabular}{l|ccc|ccc}
        \toprule[1pt]
        \multirow{2}{*}{Position} & \multicolumn{3}{c|}{AP$_{.50}$} & \multicolumn{3}{c}{AP} \\
        \cline{2-7}
        & Easy & Medium & Hard & Easy & Medium & Hard \\
        \midrule[0.5pt]
        Baseline & $95.1$ & $94.0$ & $87.2$ & $61.9$ & $59.2$ & $46.5$ \\
        Before & $95.2$ & $94.3$ & $87.5$ & $62.4$ & $60.1$ & $46.9$ \\
        Middle & $\mathbf{95.7}$ & $\mathbf{95.0}$ & $\mathbf{88.6}$ & $\mathbf{62.9}$ & $\mathbf{60.5}$ & $\mathbf{47.8}$ \\
        After & $95.3$ & $94.4$ & $88.0$ & $62.4$ & $60.2$ & $47.3$ \\
        \bottomrule[1pt]
    \end{tabular}
    \caption{The effect of the position of AutoFE and AutoFA.}
    \label{tab:different_position}
\end{table}

\begin{figure}[!t]
    \centering
    \includegraphics[width=0.7\linewidth]{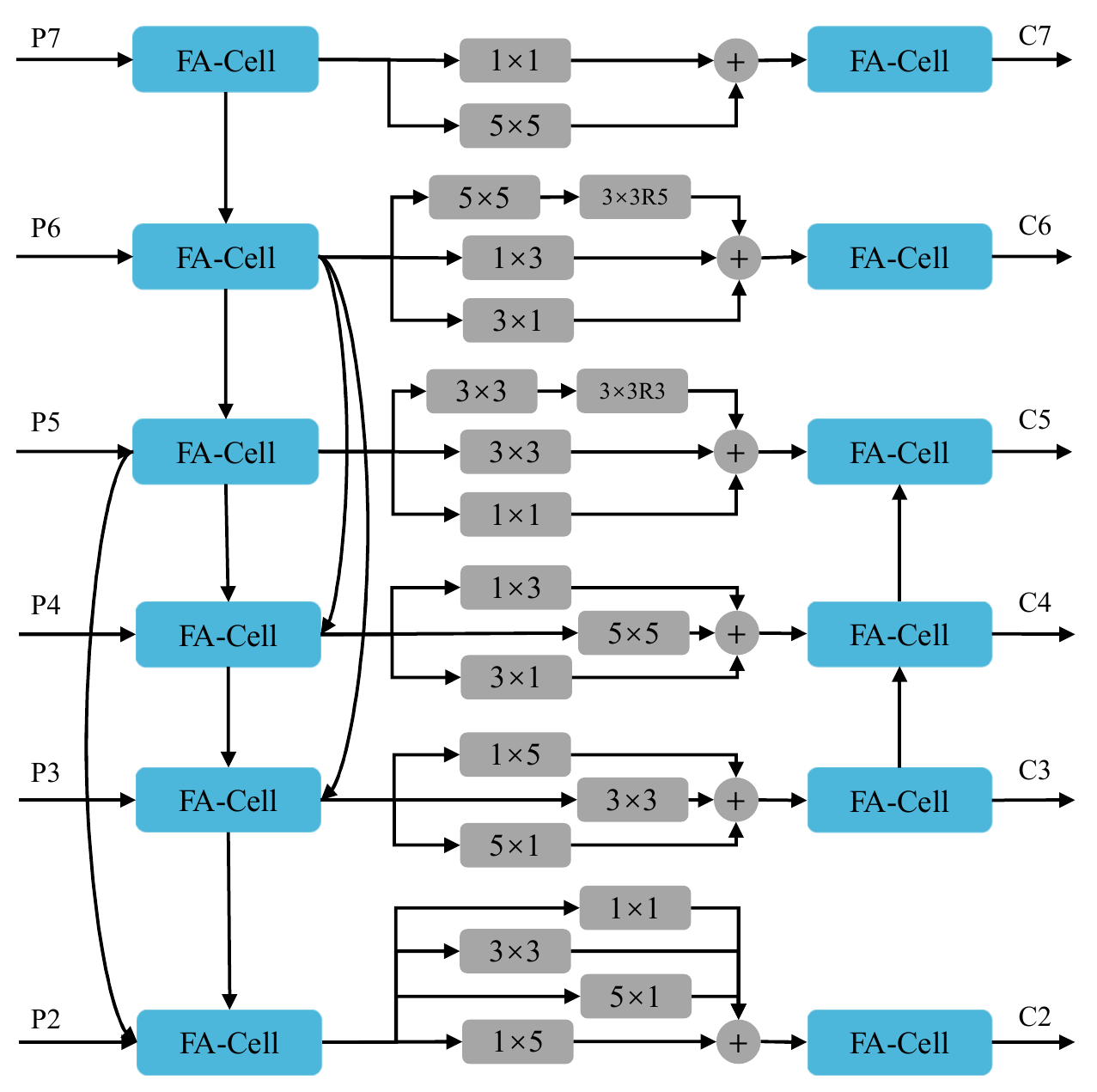}
    \vspace{-2.5mm}
    \caption{The architecture of the discovered AutoFAE, in which FA-Cell is the basic cell indicated by Fig.~\ref{fig:fpn_uint}, $m\!\times\! n$ denotes the convolution kernel size, and R$x$ is the dilated rate.}
    \vspace{-3.5mm}
    \label{fig:autofae}
\end{figure}

\begin{table}[!t]
    \centering
    \resizebox{1\linewidth}{!}{
    \begin{tabular}{l|l|ccc|ccc|c}
        \toprule[1pt]
         \multirow{2}{*}{Model} & \multirow{2}{*}{Single Path} & \multicolumn{3}{c|}{AP$_{.50}$} & \multicolumn{3}{c|}{AP} & \multirow{2}{*}{Lat.} \\
         &  & Easy & Medium & Hard & Easy & Medium & Hard &   \\
        \midrule[0.5pt]
        D$0$ & R$18$-FA-H$\times1$-$64$ & $95.7$ & $94.8$ & $88.0$ & $63.7$ & $61.1$ & $48.3$ & $3.1$ \\
        D$1$ & R$18$-FA-H$\times3$-$128$ & $96.1$ & $95.2$ & $88.8$ & $64.1$ & $61.5$ & $48.9$ & $5.7$ \\
        D$2$ & R$34$-FA-H$\times3$-$192$ & $96.4$ & $95.6$ & $89.5$ & $64.6$ & $62.3$ & $49.6$ & $10.5$ \\
        D$3$ & R$50$-FAE-H$\times3$-$192$ & $96.6$ & $95.9$ & $90.5$ & $65.1$ & $62.8$ & $50.4$ & $16.6$ \\
        D$4$ & R$50$-FAE-H$\times4$-$256$ & $97.0$ & $96.3$ & $91.2$ & $65.8$ & $63.3$ & $50.9$ & $26.2$ \\
        D$5$ & R$101$-FAE-H$\times4$-$256$ & $97.0$ & $96.5$ & $91.9$ & $65.9$ & $63.3$ & $51.7$ & $29.6$ \\
        D$6$ & R$101$-FAE-FA-FA-H$\times4$-$256$ & $97.2$ & $96.5$ & $92.5$ & $66.2$ & $63.5$ & $52.3$ & $36.1$ \\
        \bottomrule[1pt]
    \end{tabular}}
    \caption{The family of ASFD, where latency (ms) is measured with VGA-resolution images and on Nvidia V100 GPU.}
    \label{tab:trade_off}
\end{table}

\begin{figure}[!t]
    \centering
    \subfigure[WIDER Face: Hard Val]{
        \includegraphics[width=0.47\linewidth,height=0.35\linewidth]{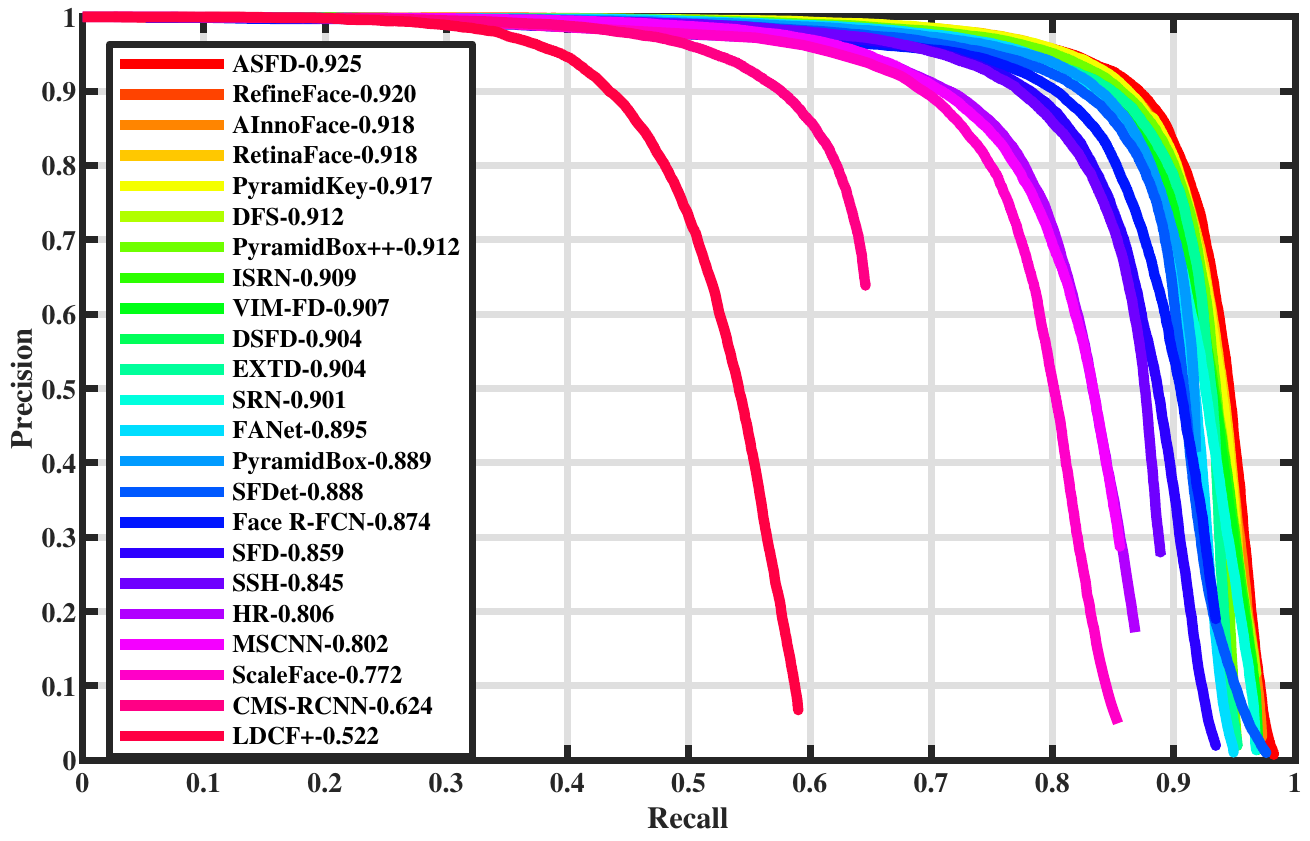}
    }
    \subfigure[WIDER Face: Hard Test]{
        \includegraphics[width=0.47\linewidth,height=0.35\linewidth]{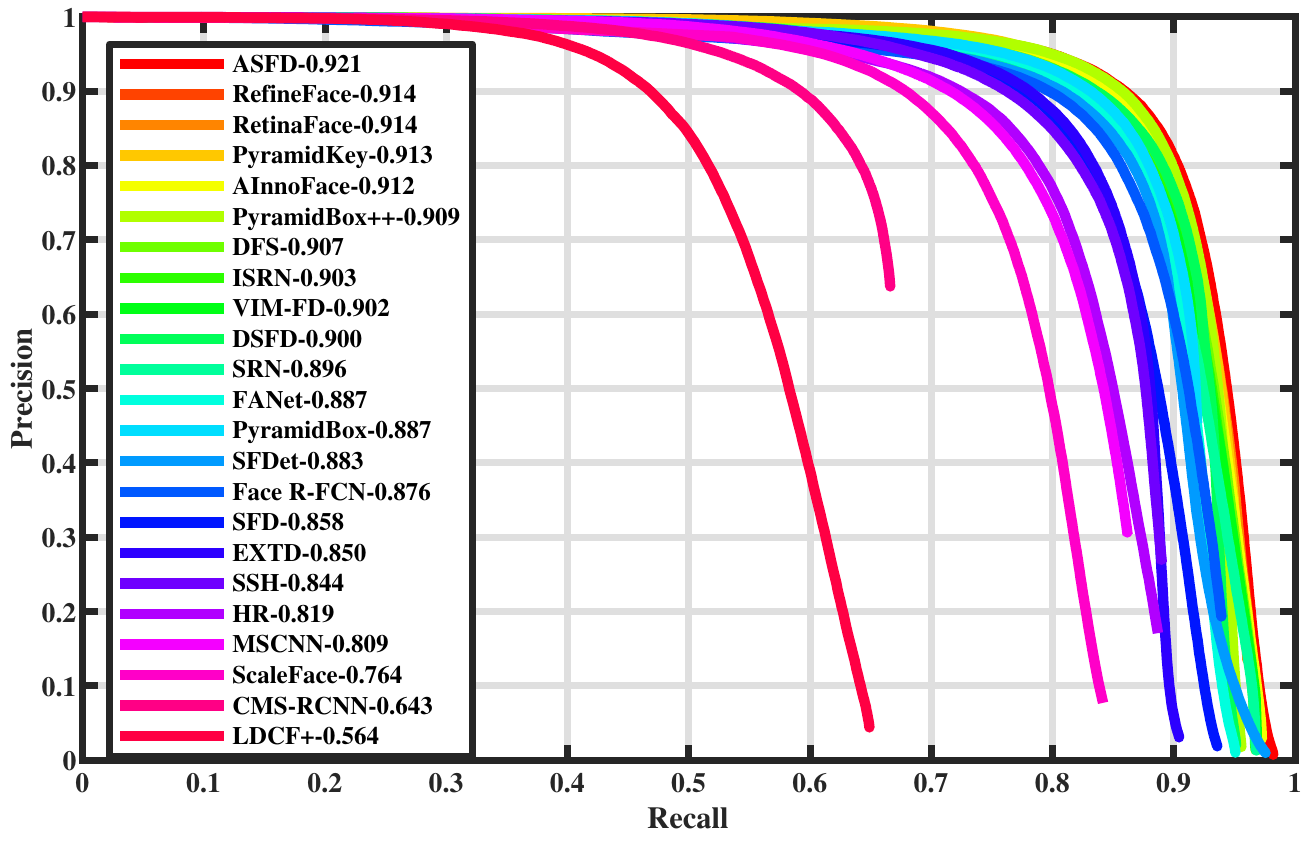}
    }
    \\
    \subfigure[FDDB: Discontinuous]{
        \includegraphics[width=0.47\linewidth,height=0.35\linewidth]{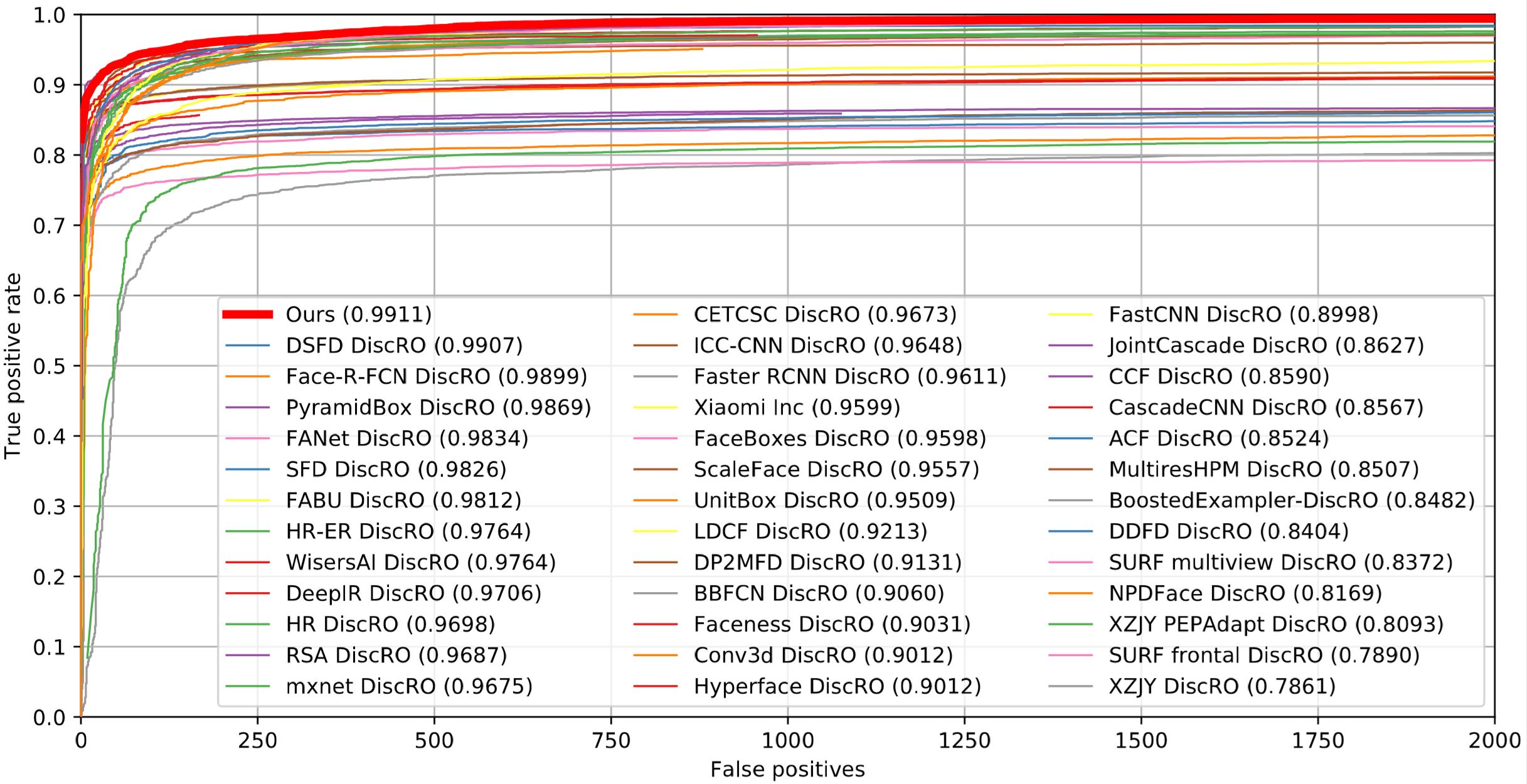}
    }
    \subfigure[FDDB: Continuous]{
        \includegraphics[width=0.47\linewidth,height=0.35\linewidth]{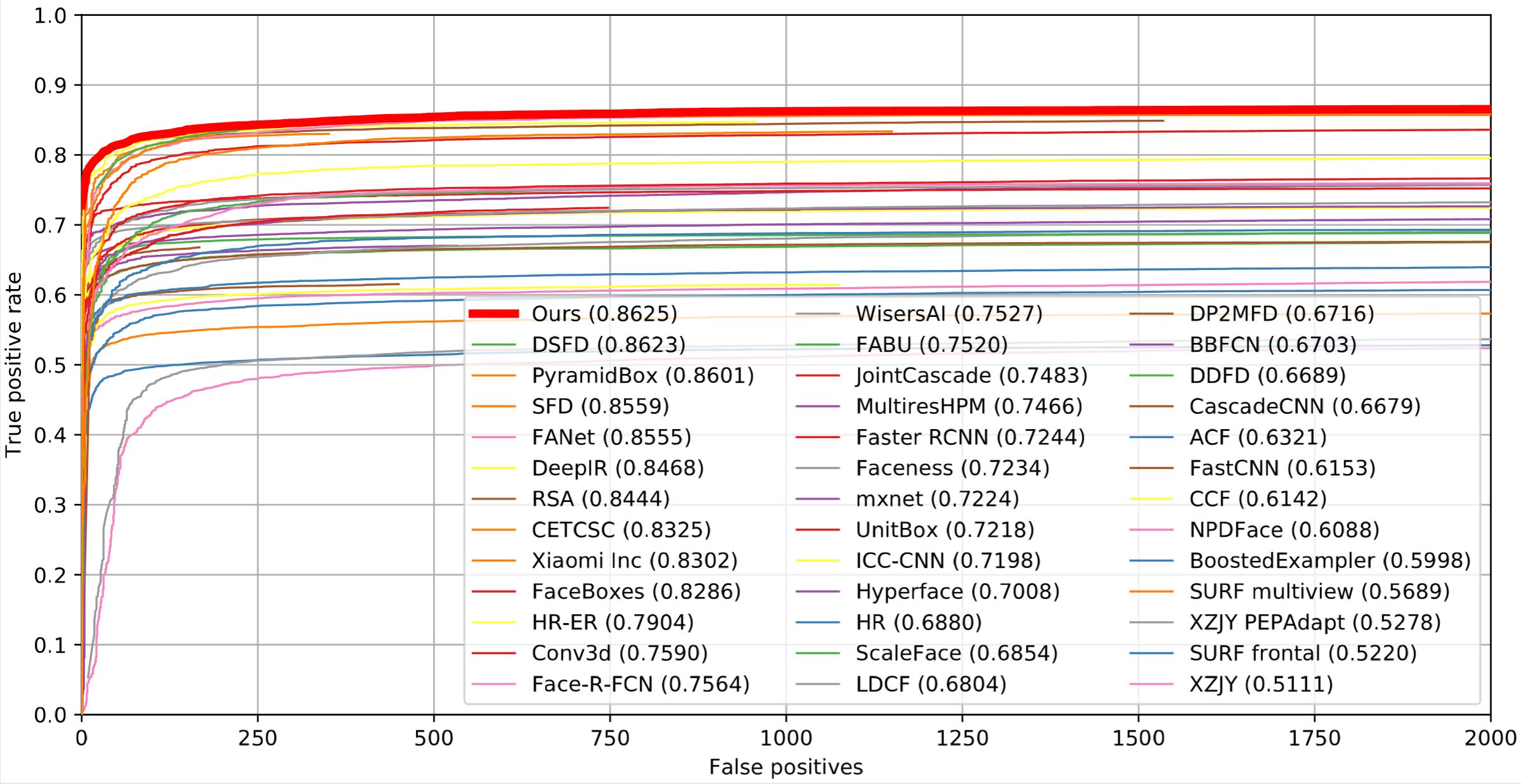}
    }
    \vspace{-2.5mm}
    \caption{Evaluation on the popular benchmarks of ASFD.}
    \vspace{-3.5mm}
    \label{fig:benchmarks}
\end{figure}

\begin{figure*}[!t]
    \centering
    \includegraphics[width=0.3\linewidth]{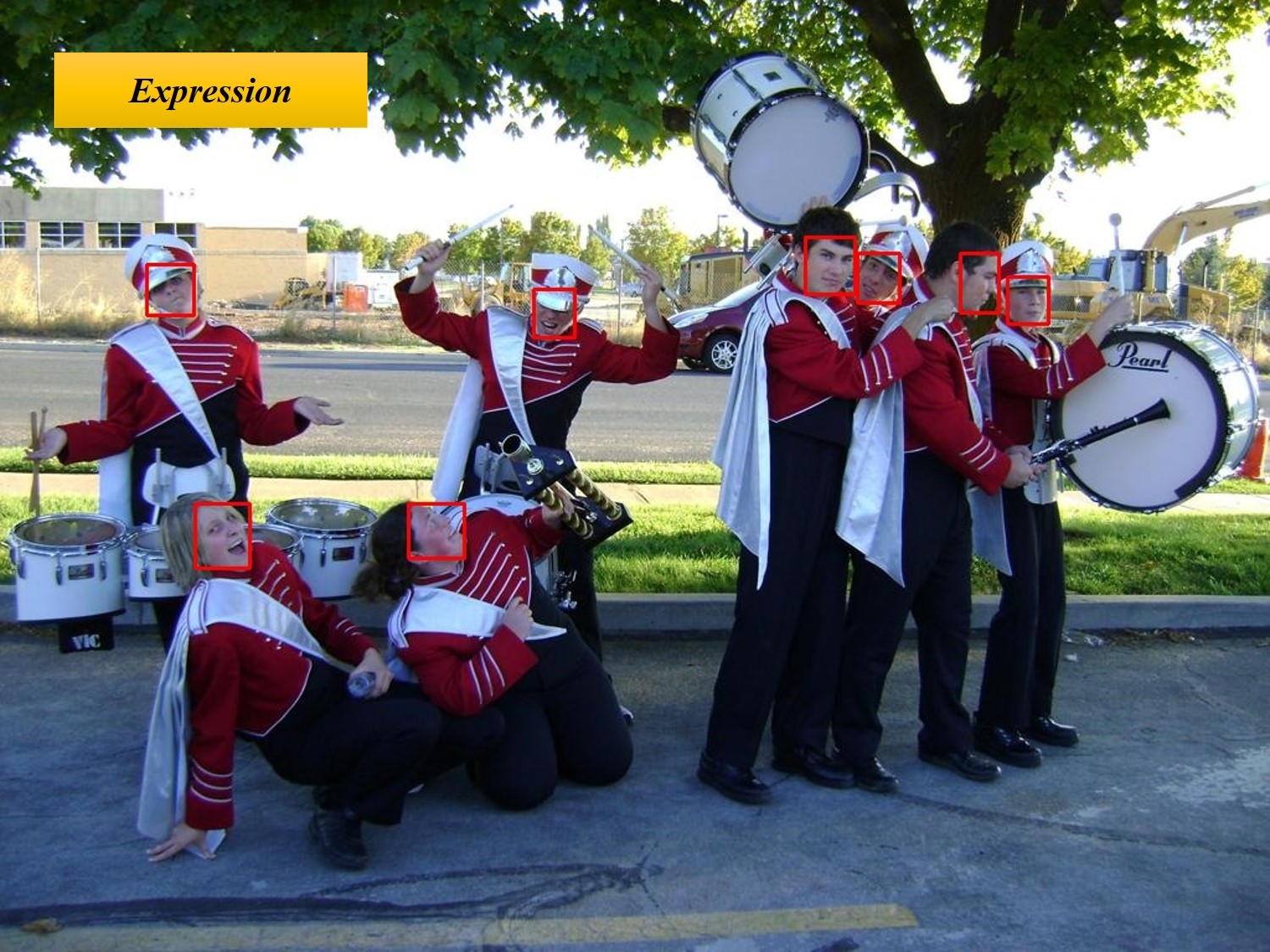}
    \includegraphics[width=0.3\linewidth]{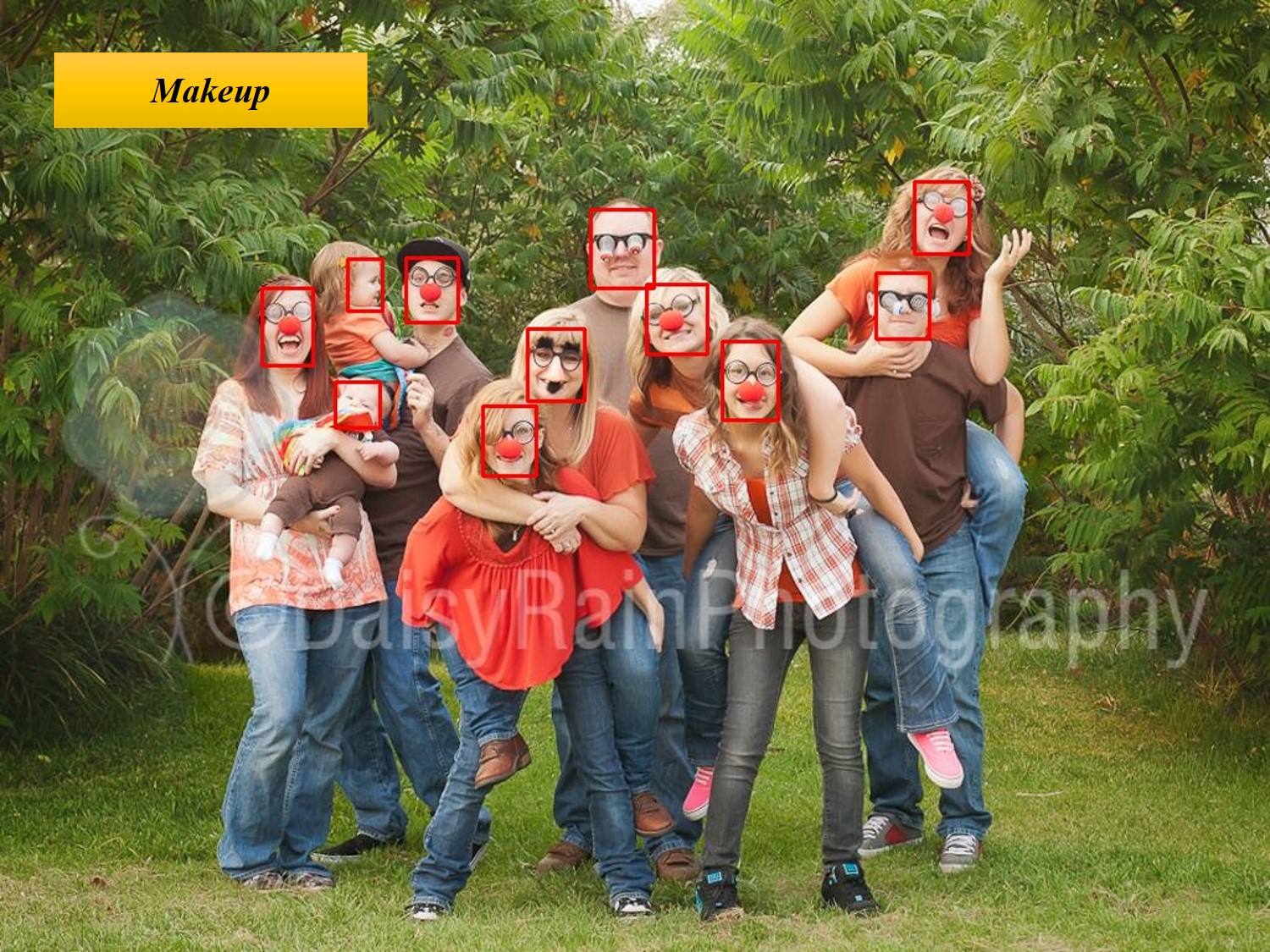}
    \includegraphics[width=0.3\linewidth]{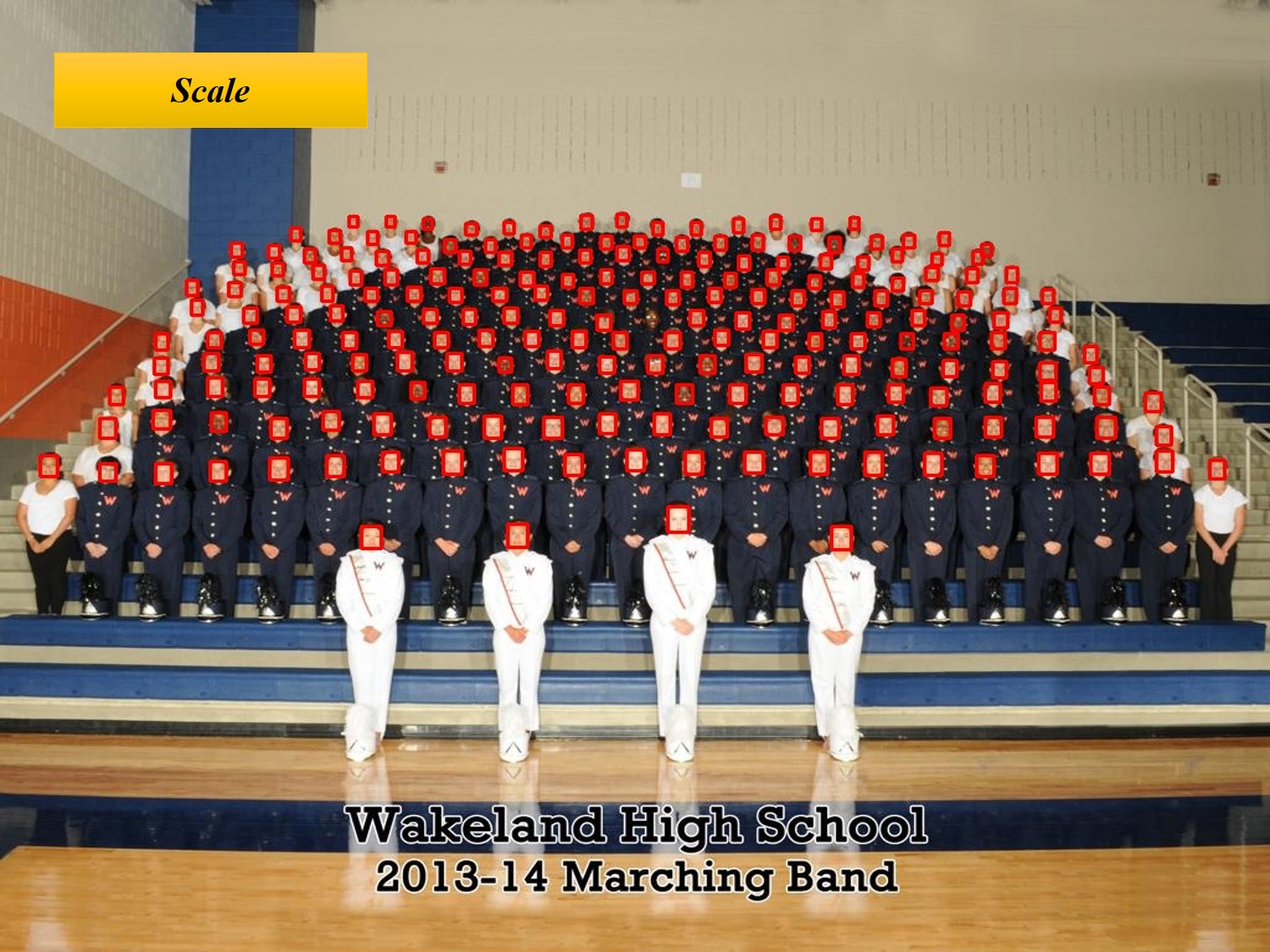}
    \vfill
    \includegraphics[width=0.3\linewidth]{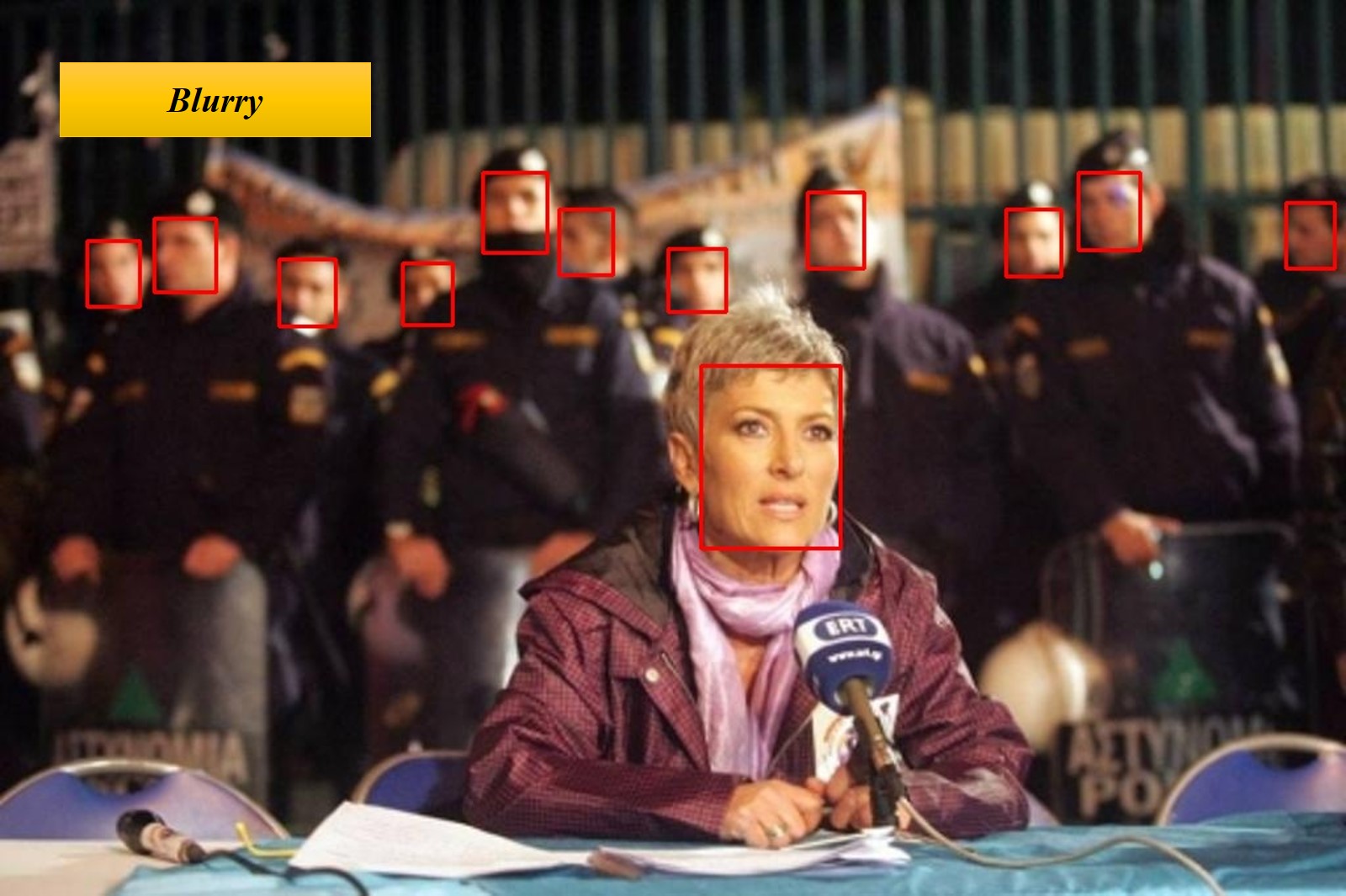}
    \includegraphics[width=0.3\linewidth]{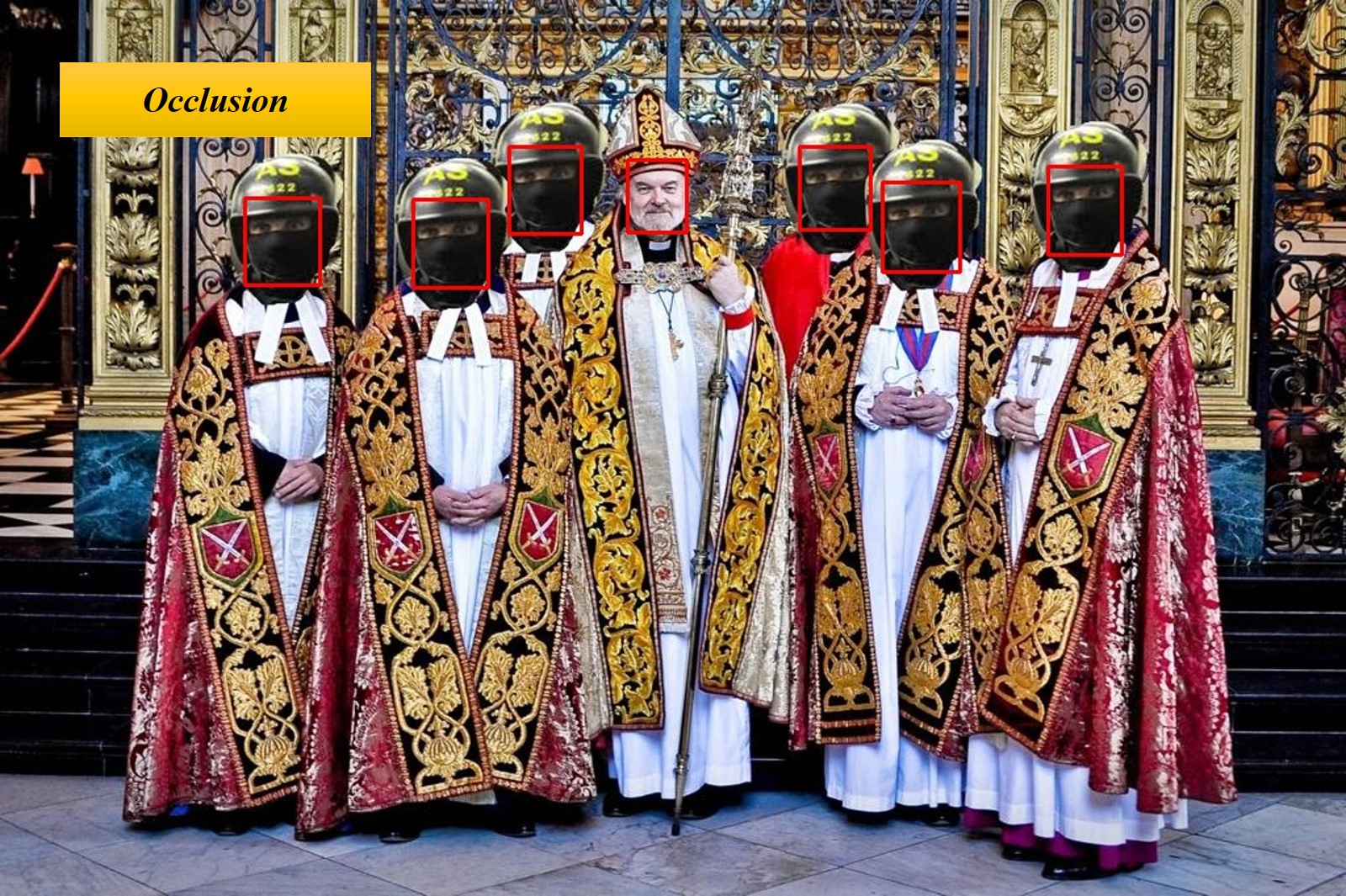}
    \includegraphics[width=0.3\linewidth]{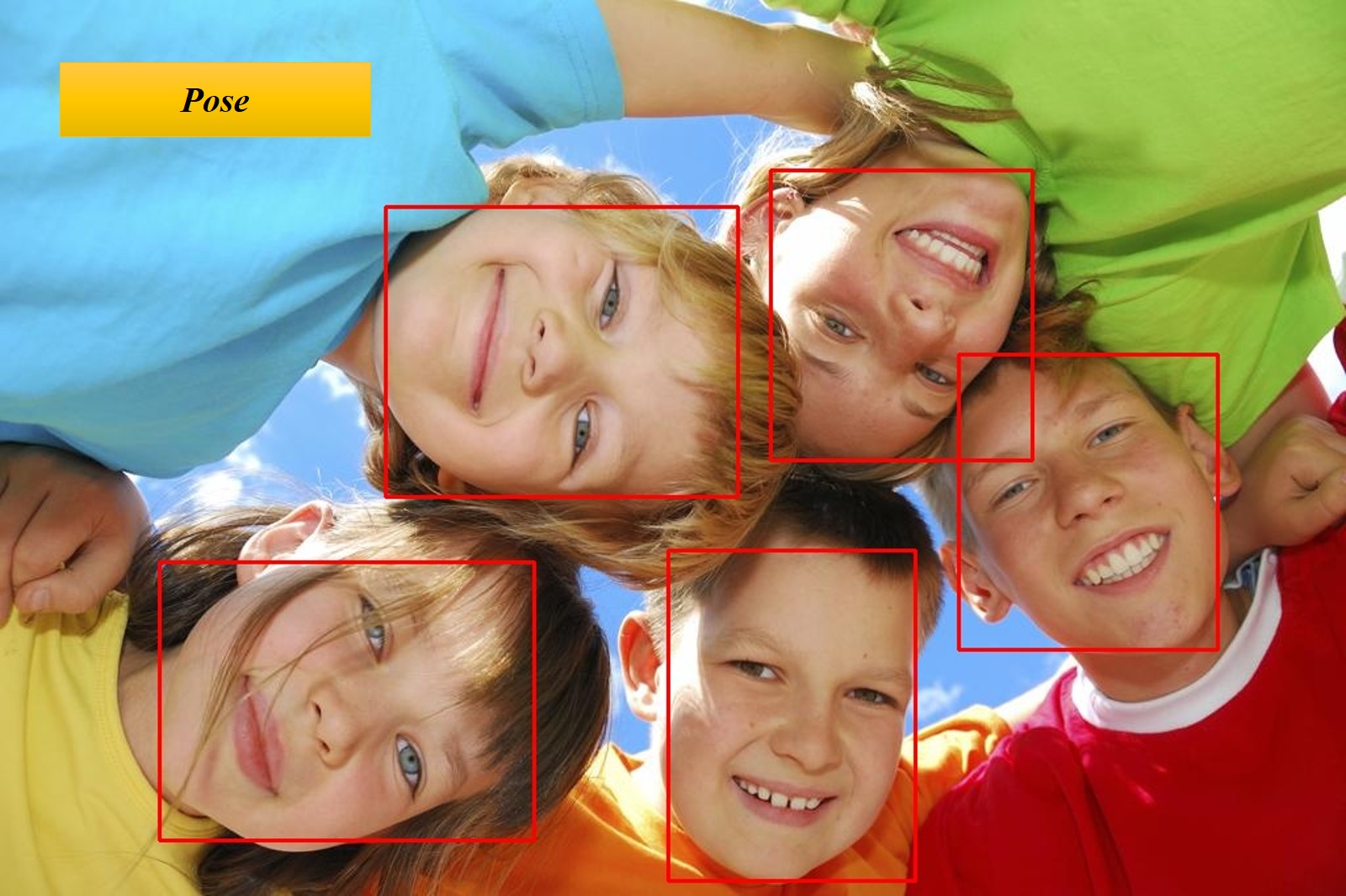}
    \vfill
    \includegraphics[width=0.3\linewidth]{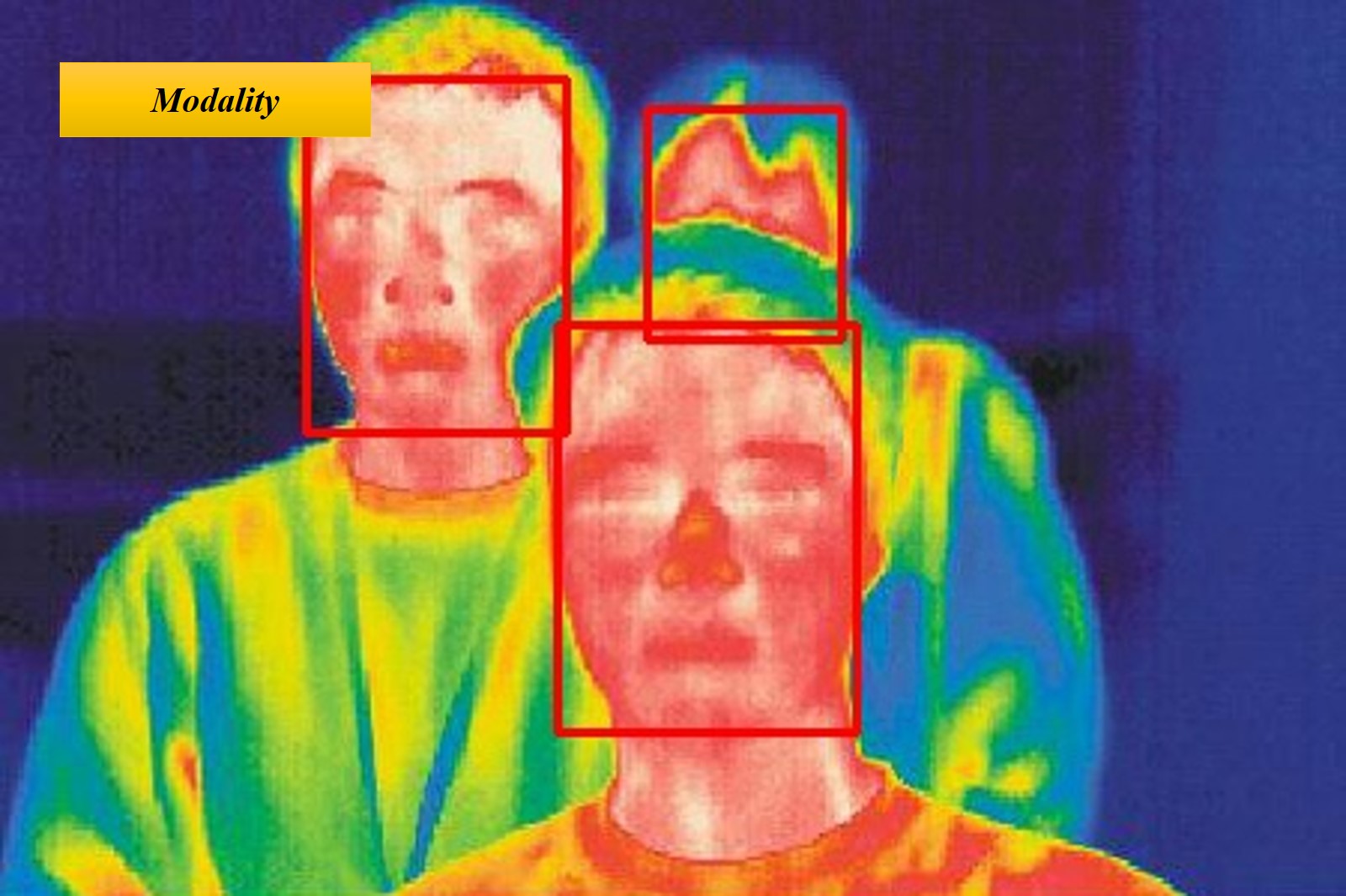}
    \includegraphics[width=0.3\linewidth]{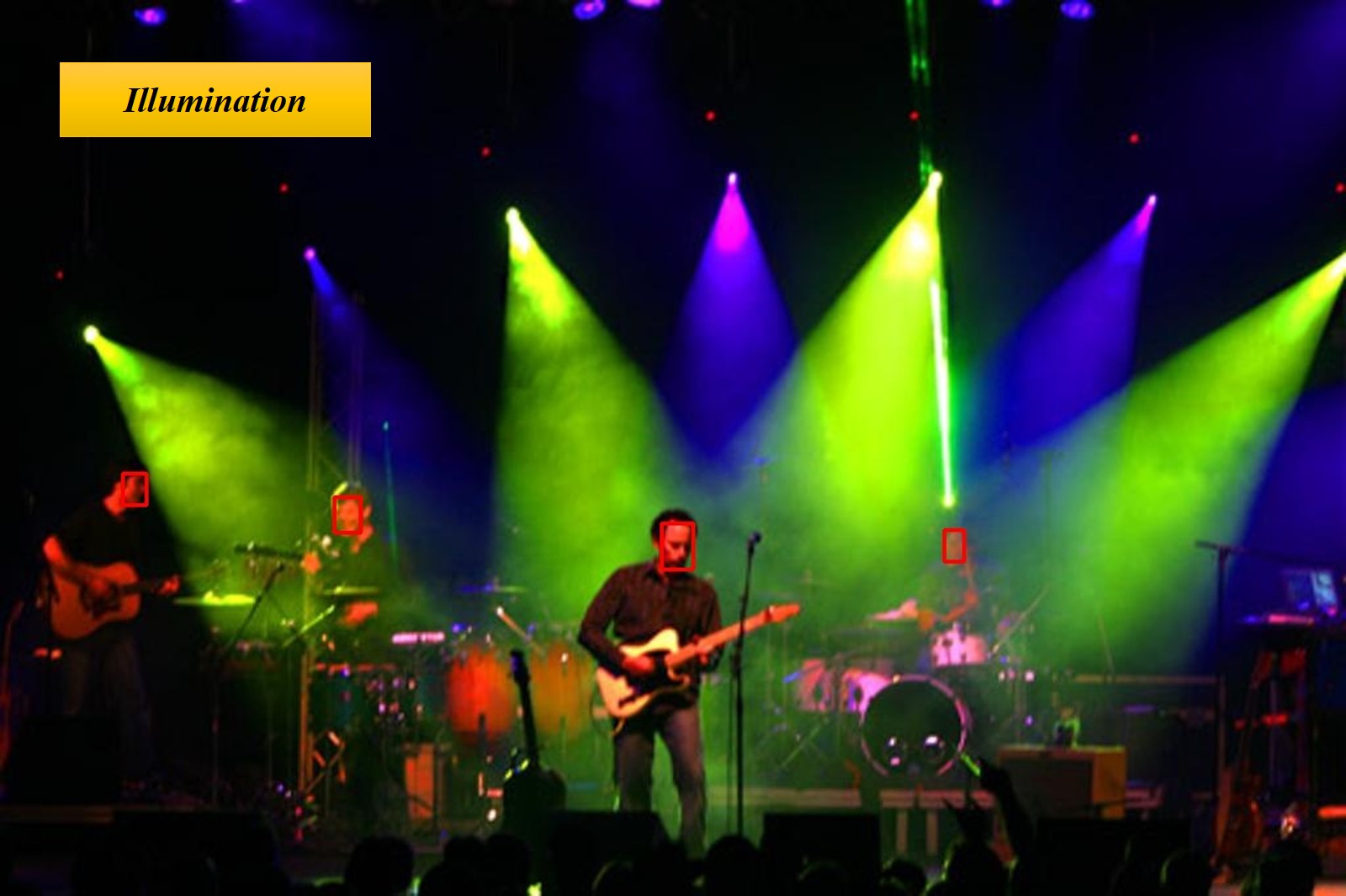}
    \includegraphics[width=0.3\linewidth]{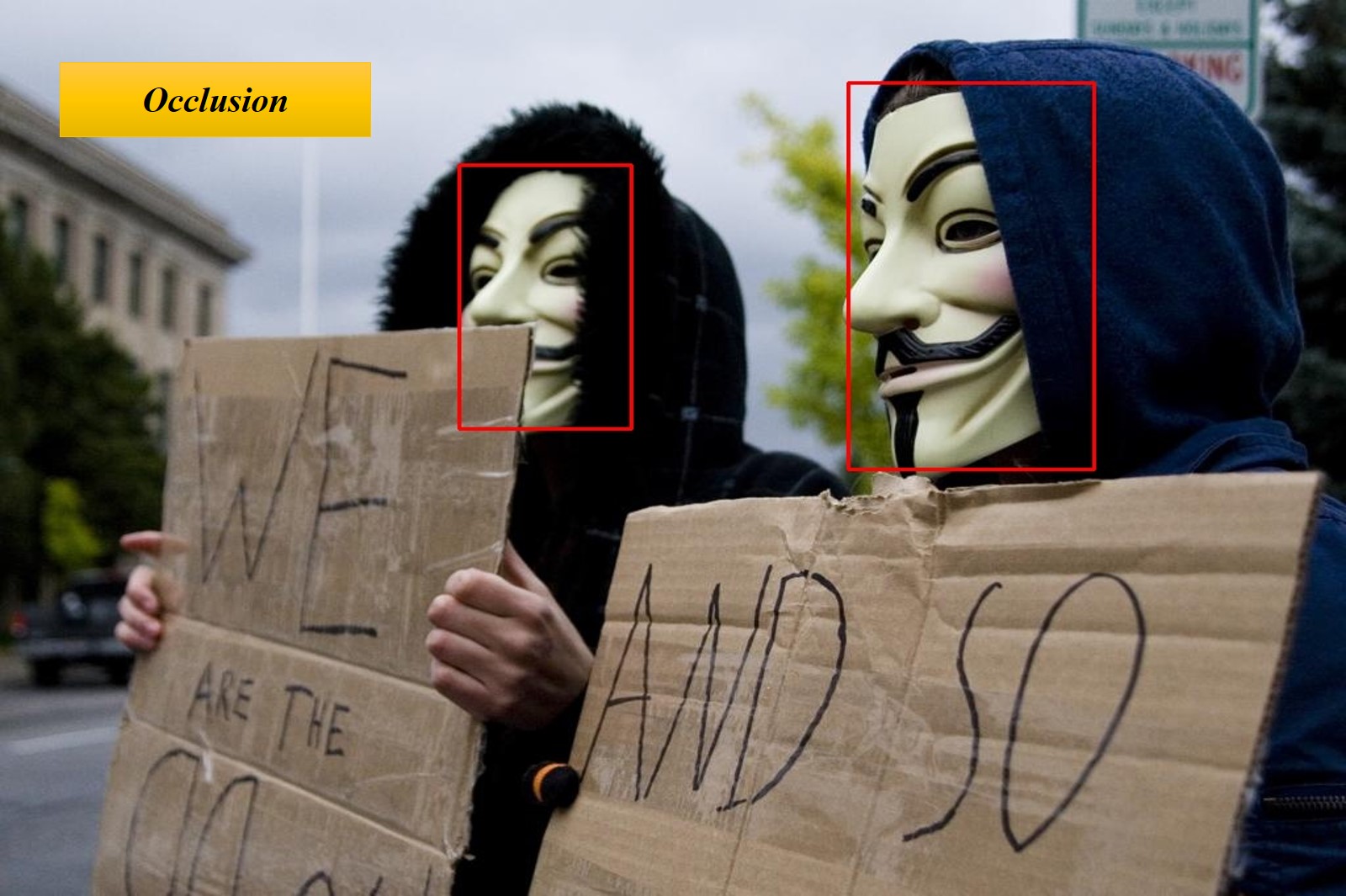}
    \vfill
    \vspace{-2.5mm}
    \caption{Illustration of ASFD to various large variations. Red bounding boxes indicate the detection confidence is above $0.8$.}
    \vspace{-3mm}
    \label{fig:visual_demo}
\end{figure*}

\subsection{Analysis on AutoFAE}
We visualize the architecture of AutoFAE in Fig.~\ref{fig:autofae}, which can match the previous conclusions drawn in the problem analysis section perfectly. 
In general, the AutoFA module aggregates pyramid features along with the sparse cross-scale and similar-scale connections instead of a fully connected manner like \cite{xu2019autofpn,wang2019nasfcos}, which avoids the performance degradation caused by large scale differences. 
Most of these cross-scale connections appear on the top-down path of AutoFA, in which the shallow features that lack semantic information are aggregated with not only the adjacent layer but also the others with rich context. 
Besides, AutoFE modules with different operations and topological structures are found for different pyramid layers. Particularly, dilated convolutions only appear in the later levels for enlarging the receptive fields, the others are almost rectangle convolutions for more diverse features. Thus, the large faces are located, and small faces in occlusion and with extreme-poses are well distinguished.
% and thus better handle the faces in occlusion or with extreme-poses.

\subsection{Model Scaling}
Next, the supernet is trained on the basis of the final AutoFAE and backbone networks of ResNet series \cite{he2016resnet}. Then, the genetic algorithm is adopted to search the single-path networks with $50$ populations and $50$ iterations. We discover $7$ single-path networks under the different GPU inference latencies, \textit{e.g.} $5$ms, $10$ms and so on. These networks are trained for $150$ epochs with the commonly used pyramid anchors \cite{tang2018pyramidbox}, and multi-scale test is employed with factors $0.5,1.0,1.5,2.0$. The detailed results are presented in Table~\ref{tab:trade_off}, in which the network architecture is indicated by single path. For instance, ``R$101$-FAE-FA-FA-H$\times4$-$256$'' means ResNet101 is adopted as the backbone, AutoFAE modules are stacked $3$ times and AutoFE module is skipped within the last two modules, convolution layers are repeated $4$ times in prediction head, and the feature channel is $256$. Obviously, our ASFD family makes a better trade-off between performance and efficiency by scaling the components and channels, especially the ASFD-D$0$ costs about $3.1$ ms, \textit{i.e.} more than 320 FPS.

\subsection{Evaluation on Benchmarks}
We evaluate our ASFD-D$6$ on the popular benchmarks, \textit{i.e.} WIDER Face \cite{yang2016wider} and FDDB \cite{jain2010fddb}, which is trained only on the training set of WIDER Face and test on these benchmarks without any fine-tuning. 
Our ASFD-D$6$ obtains the highest AP$_{.50}$ scores with $97.2/96.5/92.5$ on WIDER Face validation, $96.7/96.2/92.1$ on WIDER Face test, and $99.11$ and $86.25$ on FDDB discontinuous and continuous curves, outperforming the prior competitors by a considerable margin and setting a new state-of-the-art face detector, shown as Fig.~\ref{fig:benchmarks} (Easy and Medium results of WIDER Face are ignored due to the space limitation). More examples of our ASFD on handling face with various variations are shown in Fig.~\ref{fig:visual_demo} to demonstrate its effectiveness.
% Fig.~\ref{fig:benchmarks} shows that our ASFD-D$6$ outperforms the prior competitors by a considerable margin on both WIDER Face and FDDB, setting a new state-of-the-art face detector.

\subsection{Generalization on Generic Object Detection}
To demonstrate the generalization ability of our AutoFAE module, we evaluate the final AutoFAE module with three typical detectors, RetinaNet \cite{lin2017focal}, FCOS \cite{tian2019fcos} and Faster RCNN \cite{ren2015faster} on COCO. In particular, the original FPN module is replaced with our AutoFAE by connecting the corresponding pyramid layers, as presented in Table~\ref{tab:object_detection}, our AutoFAE module can consistently adapt to the general object domain and different detectors, with AP improvements from $0.5$ to $1.0$ points.
% Finally, the generalization of our AutoFAE is verified by plugged into different kinds of generic object detectors \cite{tian2019fcos,ren2015faster,tian2019fcos}.
% As shown in Table~\ref{tab:object_detection}, the superior performance is also obtained by replacing the FPN with our final AutoFAE.
\begin{table}[!t]
    \centering
    \resizebox{0.99\linewidth}{!}{
    \begin{tabular}{l|ccc|ccc}
        \bottomrule[1pt]
        \multirow{2}{*}{Model} & \multicolumn{3}{c|}{FPN} & \multicolumn{3}{c}{AutoFAE} \\
        \cline{2-7}
        & AP & AP$_{.50}$ & AP$_{.75}$ & AP & AP$_{.50}$ & AP$_{.75}$ \\
        \midrule[0.5pt]
        RetinaNet & $36.5$ & $55.1$ & $39.0$ & $\mathbf{37.5}$ & $\mathbf{56.6}$ & $\mathbf{39.9}$ \\
        FCOS & $38.6$ & $57.2$ & $41.7$ & $\mathbf{39.2}$ & $\mathbf{57.6}$ & $\mathbf{42.2}$ \\
        Faster RCNN & $37.4$ & $58.1$ & $40.4$ & $\mathbf{37.9}$ & $\mathbf{58.3}$ & $\mathbf{41.2}$ \\
        \bottomrule[1pt]
    \end{tabular}}
    \caption{The generalization of AutoFAE on generic object detection dataset \textit{i.e.} COCO.}
    \label{tab:object_detection}
\end{table}

\section{Conclusion}
Neural architecture search has demonstrated its successes in generic object detection about feature aggregation and enhancement. However, they cannot adapt to the domain difference between face and generic object detection and cause severe performance drops when applied to the face domain. In this paper, we analyze the reason for this phenomenon occurs and propose a face-suitable search space for feature aggregation and enhancement modules. And a better FAE module termed as AutoFAE is discovered using bi-level optimization, which outperforms the current state-of-the-art FAE modules in face detection and can be generalized to general object tasks. Finally, we automatically obtain a family of detectors with different complexities based on a supernet that achieves a better performance-efficiency trade-off.

\bibliographystyle{ACM-Reference-Format}
\balance
\bibliography{sample-base}

\end{document}